%% file: paper.tex
\pdfoutput=1
\documentclass{article}

\usepackage[preprint]{neurips_2025}

\usepackage[utf8]{inputenc} %
\usepackage[T1]{fontenc}    %
\usepackage{hyperref}       %
\usepackage{url}            %
\usepackage{booktabs}       %
\usepackage{amsfonts}       %
\usepackage{nicefrac}       %
\usepackage{microtype}      %
\usepackage{xcolor}         %
\usepackage{adjustbox}
\usepackage{ulem}
\usepackage[numbers]{natbib}

\definecolor{blue1}{RGB}{165,194,227}
\definecolor{red1}{RGB}{220,163,159}
\definecolor{green1}{RGB}{75,147,58}

\input{resources/preamble.tex}

\title{Equally Critical: Samples, Targets, and Their Mappings in Datasets}

\author{Runkang Yang$^{1}$ \quad Peng Sun$^{2,3}$ \quad Xinyi Shang$^{4}$ \quad Yi Tang$^{5}$ \quad Tao Lin$^{3,}\thanks{Corresponding author.} $ \\
$^{1}$ShanghaiTech University \quad $^{2}$Zhejiang University \quad $^{3}$Westlake University \\
$^{4}$University College London \quad ${^5}$South China University of Technology \\
}

\begin{document}

\maketitle

\input{resources/main.tex}

\bibliography{resources/reference}

\appendix

\input{resources/appendix.tex}

\newpage

\input{resources/checklist.tex}
\end{document}

%% file: resources/preamble.tex
\usepackage{amsmath,amssymb,amsthm}

\renewcommand{\gg}{\mathbf{g}}

\providecommand{\hh}{\mathbf{h}}

\providecommand{\xx}{\mathbf{x}}
\providecommand{\yy}{\mathbf{y}}

\providecommand{\mphi}{\boldsymbol{\phi}}

\providecommand{\mtheta}{\boldsymbol{\theta}}

\providecommand{\cL}{\mathcal{L}}

\providecommand{\cX}{\mathcal{X}}
\providecommand{\cY}{\mathcal{Y}}

\providecommand{\mtheta}{\boldsymbol{\theta}}

\providecommand{\mphi}{\boldsymbol{\phi}}

\providecommand{\mtheta}{\boldsymbol{\theta}}

\providecommand{\mpsi}{\boldsymbol{\psi}}

\newenvironment{talign*}
{\csname align*\endcsname}
{\endalign}

\usepackage[utf8]{inputenc}         %
\usepackage[T1]{fontenc}            %
\usepackage{url}                    %
\usepackage{booktabs}               %
\usepackage{amsfonts}               %
\usepackage{nicefrac}               %
\usepackage{microtype}              %
\usepackage{xcolor}                 %
\usepackage{algorithm}
\usepackage{algorithmic}
\usepackage{graphicx}
\usepackage{subcaption}
\usepackage[flushleft]{threeparttable}
\usepackage{float}
\usepackage{multirow}
\usepackage{xspace}
\usepackage{natbib}
\usepackage{enumitem}
\usepackage[font=small]{caption}
\usepackage{autobreak}
\usepackage{sidecap}
\usepackage{wrapfig}
\usepackage{bbding}
\usepackage[toc, page, header]{appendix}
\usepackage{tikz}
\usepackage{xcolor}
\usepackage{pifont}
\usepackage{mdframed}
\usepackage{colortbl}

\hypersetup{
  colorlinks=true,
  linkcolor=blue,
  citecolor=blue,
  urlcolor=blue
}

\definecolor{coral}{RGB}{255,127,80}
\definecolor{darkgreen}{RGB}{0,100,0}
\definecolor{darkyellow}{RGB}{204,153,0}
\definecolor{salmon}{RGB}{250,128,114}

\newcommand{\transparentred}[1]{%
  \tikz[baseline=(X.base)] \node[fill=red, fill opacity=0.1, text opacity=1, inner sep=2pt, outer sep=0pt] (X) {#1};%
}
\newcommand{\thmref}[1]{\hyperref[#1]{\transparentred{Theorem~\ref*{#1}}}}

\newcommand{\transparentgray}[1]{%
  \tikz[baseline=(X.base)] \node[fill=gray, fill opacity=0.1, text opacity=1, inner sep=2pt, outer sep=0pt] (X) {#1};%
}
\newcommand{\defref}[1]{\hyperref[#1]{\transparentgray{Definition~\ref*{#1}}}}

\newcommand{\transparentblue}[1]{%
  \tikz[baseline=(X.base)] \node[fill=blue, fill opacity=0.1, text opacity=1, inner sep=2pt, outer sep=0pt] (X) {#1};%
}
\newcommand{\propref}[1]{\hyperref[#1]{\transparentblue{Proposition~\ref*{#1}}}}

\newcommand{\transparentgreen}[1]{%
  \tikz[baseline=(X.base)] \node[fill=green, fill opacity=0.1, text opacity=1, inner sep=2pt, outer sep=0pt] (X) {#1};%
}
\newcommand{\assumpref}[1]{\hyperref[#1]{\transparentgreen{Assumption~\ref*{#1}}}}

\newcommand{\transparentyellow}[1]{%
  \tikz[baseline=(X.base)] \node[fill=yellow, fill opacity=0.1, text opacity=1, inner sep=2pt, outer sep=0pt] (X) {#1};%
}
\newcommand{\remarkref}[1]{\hyperref[#1]{\transparentyellow{Remark~\ref*{#1}}}}

\newcommand{\transparentpurple}[1]{%
  \tikz[baseline=(X.base)] \node[fill=purple, fill opacity=0.1, text opacity=1, inner sep=2pt, outer sep=0pt] (X) {#1};%
}
\newcommand{\hypref}[1]{\hyperref[#1]{\transparentpurple{Hypothesis~\ref*{#1}}}}

\newcommand{\transparentorange}[1]{%
  \tikz[baseline=(X.base)] \node[fill=orange, fill opacity=0.1, text opacity=1, inner sep=2pt, outer sep=0pt] (X) {#1};%
}
\newcommand{\conjref}[1]{\hyperref[#1]{\transparentorange{Conjecture~\ref*{#1}}}}

\newcommand{\transparentcyan}[1]{%
  \tikz[baseline=(X.base)] \node[fill=cyan, fill opacity=0.1, text opacity=1, inner sep=2pt, outer sep=0pt] (X) {#1};%
}
\newcommand{\lemref}[1]{\hyperref[#1]{\transparentcyan{Lemma~\ref*{#1}}}}

\newcommand{\transparentmagenta}[1]{%
  \tikz[baseline=(X.base)] \node[fill=magenta, fill opacity=0.1, text opacity=1, inner sep=2pt, outer sep=0pt] (X) {#1};%
}
\newcommand{\corref}[1]{\hyperref[#1]{\transparentmagenta{Corollary~\ref*{#1}}}}

\newcommand{\transparentpink}[1]{%
  \tikz[baseline=(X.base)] \node[fill=pink, fill opacity=0.1, text opacity=1, inner sep=2pt, outer sep=0pt] (X) {#1};%
}
\newcommand{\noteref}[1]{\hyperref[#1]{\transparentpink{Notation~\ref*{#1}}}}

\newcommand{\transparentviolet}[1]{%
  \tikz[baseline=(X.base)] \node[fill=violet, fill opacity=0.1, text opacity=1, inner sep=2pt, outer sep=0pt] (X) {#1};%
}
\newcommand{\claimref}[1]{\hyperref[#1]{\transparentviolet{Claim~\ref*{#1}}}}

\newcommand{\transparentsalmon}[1]{%
  \tikz[baseline=(X.base)] \node[fill=salmon, fill opacity=0.1, text opacity=1, inner sep=2pt, outer sep=0pt] (X) {#1};%
}
\newcommand{\probref}[1]{\hyperref[#1]{\transparentsalmon{Problem~\ref*{#1}}}}

\newcommand{\transparentlavender}[1]{%
  \tikz[baseline=(X.base)] \node[fill=lavender, fill opacity=0.1, text opacity=1, inner sep=2pt, outer sep=0pt] (X) {#1};%
}
\newcommand{\obsref}[1]{\hyperref[#1]{\transparentlavender{Observation~\ref*{#1}}}}

\newcommand{\transparentlime}[1]{%
  \tikz[baseline=(X.base)] \node[fill=lime, fill opacity=0.1, text opacity=1, inner sep=2pt, outer sep=0pt] (X) {#1};%
}
\newcommand{\algref}[1]{\hyperref[#1]{\transparentlime{Algorithm~\ref*{#1}}}}

\newcommand{\transparentteal}[1]{%
  \tikz[baseline=(X.base)] \node[fill=teal, fill opacity=0.1, text opacity=1, inner sep=2pt, outer sep=0pt] (X) {#1};%
}
\newcommand{\figref}[1]{\hyperref[#1]{\transparentteal{Figure~\ref*{#1}}}}

\newcommand{\transparentdarkgreen}[1]{%
  \tikz[baseline=(X.base)] \node[fill=darkgreen, fill opacity=0.1, text opacity=1, inner sep=2pt, outer sep=0pt] (X) {#1};%
}
\newcommand{\tabref}[1]{\hyperref[#1]{\transparentdarkgreen{Table~\ref*{#1}}}}

\newcommand{\transparentdarkyellow}[1]{%
  \tikz[baseline=(X.base)] \node[fill=darkyellow, fill opacity=0.1, text opacity=1, inner sep=2pt, outer sep=0pt] (X) {#1};%
}
\newcommand{\secref}[1]{\hyperref[#1]{\transparentdarkyellow{Section~\ref*{#1}}}}

\newcommand{\transparentcoral}[1]{%
  \tikz[baseline=(X.base)] \node[fill=coral, fill opacity=0.1, text opacity=1, inner sep=2pt, outer sep=0pt] (X) {#1};%
}
\newcommand{\appref}[1]{\hyperref[#1]{\transparentcoral{Appendix~\ref*{#1}}}}

\newtheoremstyle{custom}
{1pt} %
{1pt} %
{\itshape} %
{} %
{\bfseries} %
{} %
{ } %
{\thmname{#1} \thmnumber{#2} \thmnote{(#3)} . } %

\theoremstyle{custom}

\newtheorem{innerdefinition}{Definition}
\newtheorem{innerproposition}{Proposition}
\newtheorem{innerassumption}{Assumption}
\newtheorem{innerremark}{Remark}
\newtheorem{innertheorem}{Theorem}
\newtheorem{innerhypothesis}{Hypothesis}
\newtheorem{innerconjecture}{Conjecture}
\newtheorem{innerlemma}{Lemma}
\newtheorem{innercorollary}{Corollary}
\newtheorem{innernotation}{Notation}
\newtheorem{innerclaim}{Claim}
\newtheorem{innerproblem}{Problem}

\newtheorem{innerobservation}{Observation}

\newmdenv[
  backgroundcolor=gray!10,
  linecolor=gray!100,
  linewidth=0.8pt,
  skipabove=2pt,
  skipbelow=2pt,
  innertopmargin=10pt,
  innerbottommargin=5pt,
  innerleftmargin=5pt,
  innerrightmargin=5pt,
]{definitionframe}

\newmdenv[
  backgroundcolor=blue!10,
  linecolor=blue!100,
  linewidth=0.8pt,
  skipabove=2pt,
  skipbelow=2pt,
  innertopmargin=10pt,
  innerbottommargin=5pt,
  innerleftmargin=5pt,
  innerrightmargin=5pt,
]{propositionframe}

\newmdenv[
  backgroundcolor=green!10,
  linecolor=green!100,
  linewidth=0.8pt,
  skipabove=2pt,
  skipbelow=2pt,
  innertopmargin=10pt,
  innerbottommargin=5pt,
  innerleftmargin=5pt,
  innerrightmargin=5pt,
]{assumptionframe}

\newmdenv[
  backgroundcolor=yellow!10,
  linecolor=yellow!100,
  linewidth=0.8pt,
  skipabove=2pt,
  skipbelow=2pt,
  innertopmargin=10pt,
  innerbottommargin=5pt,
  innerleftmargin=5pt,
  innerrightmargin=5pt,
]{remarkframe}

\newmdenv[
  backgroundcolor=red!10,
  linecolor=red!100,
  linewidth=0.8pt,
  skipabove=2pt,
  skipbelow=2pt,
  innertopmargin=10pt,
  innerbottommargin=5pt,
  innerleftmargin=5pt,
  innerrightmargin=5pt,
]{theoremframe}

\newmdenv[
  backgroundcolor=purple!10,
  linecolor=purple!100,
  linewidth=0.8pt,
  skipabove=2pt,
  skipbelow=2pt,
  innertopmargin=10pt,
  innerbottommargin=5pt,
  innerleftmargin=5pt,
  innerrightmargin=5pt,
]{hypothesisframe}

\newmdenv[
  backgroundcolor=orange!10,
  linecolor=orange!100,
  linewidth=0.8pt,
  skipabove=2pt,
  skipbelow=2pt,
  innertopmargin=10pt,
  innerbottommargin=5pt,
  innerleftmargin=5pt,
  innerrightmargin=5pt,
]{conjectureframe}

\newmdenv[
  backgroundcolor=cyan!10,
  linecolor=cyan!100,
  linewidth=0.8pt,
  skipabove=2pt,
  skipbelow=2pt,
  innertopmargin=10pt,
  innerbottommargin=5pt,
  innerleftmargin=5pt,
  innerrightmargin=5pt,
]{lemmaframe}

\newmdenv[
  backgroundcolor=magenta!10,
  linecolor=magenta!100,
  linewidth=0.8pt,
  skipabove=2pt,
  skipbelow=2pt,
  innertopmargin=10pt,
  innerbottommargin=5pt,
  innerleftmargin=5pt,
  innerrightmargin=5pt,
]{corollaryframe}

\newmdenv[
  backgroundcolor=pink!10,
  linecolor=pink!100,
  linewidth=0.8pt,
  skipabove=2pt,
  skipbelow=2pt,
  innertopmargin=10pt,
  innerbottommargin=5pt,
  innerleftmargin=5pt,
  innerrightmargin=5pt,
]{notationframe}

\newmdenv[
  backgroundcolor=violet!10,
  linecolor=violet!100,
  linewidth=0.8pt,
  skipabove=2pt,
  skipbelow=2pt,
  innertopmargin=10pt,
  innerbottommargin=5pt,
  innerleftmargin=5pt,
  innerrightmargin=5pt,
]{claimframe}

\newmdenv[
  backgroundcolor=salmon!10,
  linecolor=salmon!100,
  linewidth=0.8pt,
  skipabove=2pt,
  skipbelow=2pt,
  innertopmargin=10pt,
  innerbottommargin=5pt,
  innerleftmargin=5pt,
  innerrightmargin=5pt,
]{problemframe}

\newmdenv[
  backgroundcolor=lavender!10,
  linecolor=lavender!100,
  linewidth=0.8pt,
  skipabove=2pt,
  skipbelow=2pt,
  innertopmargin=10pt,
  innerbottommargin=5pt,
  innerleftmargin=5pt,
  innerrightmargin=5pt,
]{observationframe}

\newenvironment{claim}
{\begin{claimframe}\begin{innerclaim}}
      {\end{innerclaim}\end{claimframe}}

\usepackage[textwidth=2.cm,textsize=tiny]{todonotes}

\newcommand{\SA}{\textsc{Strategy A}\xspace}
\newcommand{\SB}{\textsc{Strategy B}\xspace}
\newcommand{\SC}{\textsc{Strategy C}\xspace}

\definecolor{darkgreen}{RGB}{0, 100, 0}

\newcommand{\IPC}{\texttt{IPC}\xspace}

%% file: resources/main.tex
\vspace{-15pt}

\begin{abstract}

    Data inherently possesses dual attributes: samples and targets. For targets, knowledge distillation has been widely employed to accelerate model convergence, primarily relying on teacher-generated soft target supervision. 
    Conversely, recent advancements in data-efficient learning have emphasized sample optimization techniques, such as dataset distillation, while neglected the critical role of target. This dichotomy motivates our investigation into understanding how both sample and target collectively influence training dynamic.
    To address this gap, we first establish a taxonomy of existing paradigms through the lens of sample-target interactions, categorizing them into distinct sample-to-target mapping strategies. Building upon this foundation, we then propose a novel unified loss framework to assess their impact on training efficiency.
    Through extensive empirical studies on our proposed strategies, we comprehensively analyze \textit{how variations in target and sample types, quantities, and qualities} influence model training, providing \textit{\textbf{six key insights}} to enhance training efficacy. %

\end{abstract}

\vspace{-15pt}
\section{Introduction}
\label{sec:introduction}

\begin{wrapfigure}{r}{0.50\textwidth}
    \centering
    \vspace{-1em}
    \includegraphics[width=0.9\linewidth]{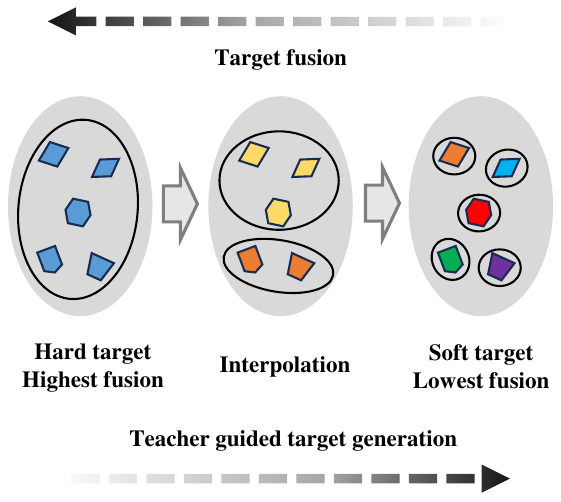}
    \caption{\small
        The key difference between hard and soft targets lies in \textbf{\textit{target fusion}}: hard targets map all \textit{(augmented)} samples of the same class to a shared target, while soft targets assign each one a unique target.
    }
    \label{fig:fusion}
    \vspace{-1.em}
\end{wrapfigure}

The remarkable progress of deep learning has been largely propelled by the availability of massive amounts of data \citep{deng2009imagenet, hestness2017deep, kaplan2020scaling}. 
Yet, the question of how to effectively utilize data to maximize model training efficiency still remains a fundamental challenge \citep{brown2020language, cheng2017survey}.
As the cornerstone of model training, data inherently possesses two fundamental attributes: samples and targets \footnote{
    In this paper, we decouple dataset into two essential components: samples $X$ and targets $Y$.
    For example, in a classification dataset, $X$ represents the images, while $Y$ consists of the one-hot encoded targets.
}. 
Among these, knowledge distillation \citep{hinton2015distilling} has been proposed to effectively accelerate model convergence using soft targets generated by pre-trained teachers, and has gained huge success to drive the unprecedented development of lightweight models like DistilBERT \citep{sanh2019distilbert} and DeepSeek-R1 Series \citep{guo2025deepseek}. %
Sample optimization, on the other hand, has been a central focus of data-efficient learning \citep{zhu2023data,ray2025molecular} driven by the increasing demand for training models under limited data.
Recent studies \citep{sorscher2022beyond,fan2024scaling} have explore strategies that can leverage smaller datasets without compromising performance, like data pruning strategies \citep{paul2021deep} can achieve competitive performance with reduced training budgets by focusing on the most informative samples.
Optimization on samples also include training using synthetic \citep{fan2024scaling} and augmented \citep{geiping2022much} images.

Despite these achievements, \textit{prior research have predominantly treated samples and targets as independent entities}.
Advancements in data-efficient learning methods \citep{sun2024diversity, yin2023squeeze, guo2023towards} have highlighted the importance of both samples and targets.
For instance, efficient dataset distillation methods heavily rely on soft targets generated by pre-trained models \citep{sun2024diversity}.%
Furthermore, \citep{sun2024efficiency} provide theoretical insights demonstrating that both targets and samples within a dataset significantly influence training dynamics and convergence.
A detailed discussion of related work can be found in \appref{app:related}.

In this work, we demonstrate that in supervised learning,
an implicit and inherent correspondence exists between samples and their associated targets,
characterized by the degree of target fusion shared among different samples.
Hard targets indicate that all samples within the same class share an identical target,
while soft targets represent the opposite, as illustrated in~\figref{fig:fusion}.
However, hard and soft targets merely represent two cases of this relationship, and how this correspondence affects model training still remain underexplored.

To enable a systematic examination of samples and targets for model  training, we first review existing paradigms in the sample-target relationship, categorizing them into distinct sample-to-target mapping strategies:
multiple-to-one mappings (typical in conventional supervised learning), multiple-to-multiple mappings (common in knowledge distillation \cite{hinton2015distilling} using teacher models), and the proposed multiple-to-few mappings, as detailed in \secref{sec:definition_strategy},
we then propose a unified loss framework to evaluate their impact on model training in \secref{sec:design_loss}.
Building on this foundation, we take computer vision task as a case study and conduct a comprehensive analysis of \textit{how variations in target and sample types, quantities, and qualities affect training efficiency.}
Our analysis covers: (a) diverse target generation strategies in \secref{sec:exp1} \uline{(target quantity)},
(b) teacher models with different performance in \secref{sec:exp2} and various augmentation in \secref{sec:exp3} \uline{(target quality)},
and (c) \uline{sample quantity and quality} in \secref{sec:sample}.
We list our \textbf{six key findings below}:
\begin{enumerate}[label=(\alph*),leftmargin=16pt, itemsep=0pt]
    \item Under the same training budge, decoupling backbone and classifier for separate training can greatly benefit downstream tasks. %
    \item Stronger teacher models enhance final performance, but weak ones aid early learning.
    \item Our proposed \SC can achieve higher final accuracy compared to traditional knowledge distillation, while retaining the ability to accelerate early-stage training.
    \item Teacher models trained with augmentations like MixUp do not consistently improve student performance, while RandomResizedCrop adversely affects student training.
    \item With limited data, soft targets provide advantages, but larger sample sizes favor one-hot targets.
    \item RandomResizedCrop is most effective for one-hot targets, while Mixed-based augmentation excels with soft targets.
\end{enumerate}

\begin{figure*}[t]
    \centering
    \begin{subfigure}{.32\textwidth}  %
        \centering
        \includegraphics[width=1.0\linewidth]{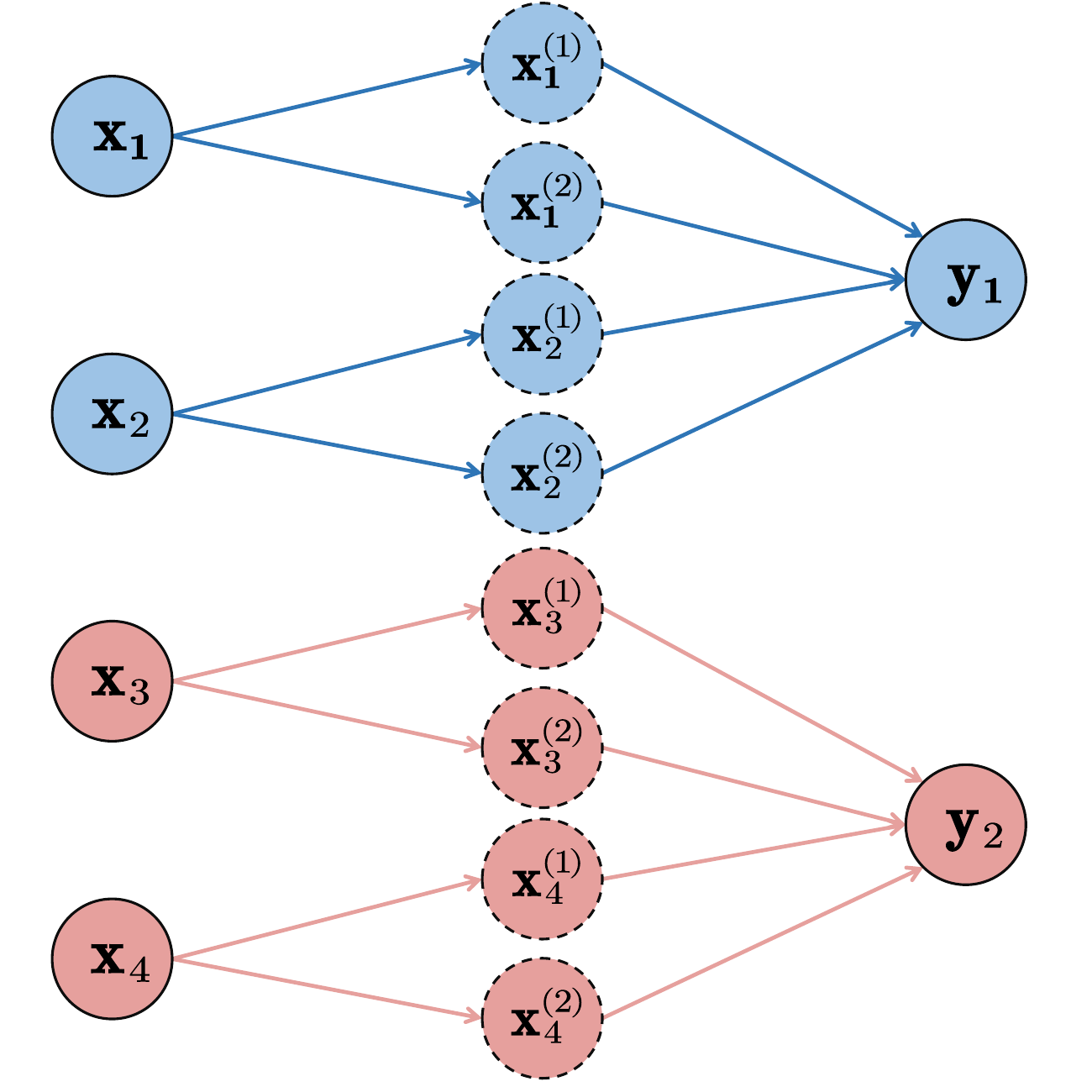}
        \caption{\SA}
        \label{fig:sa}
    \end{subfigure}
    \hfill %
    \begin{subfigure}{.32\textwidth}
        \centering
        \includegraphics[width=1.0\linewidth]{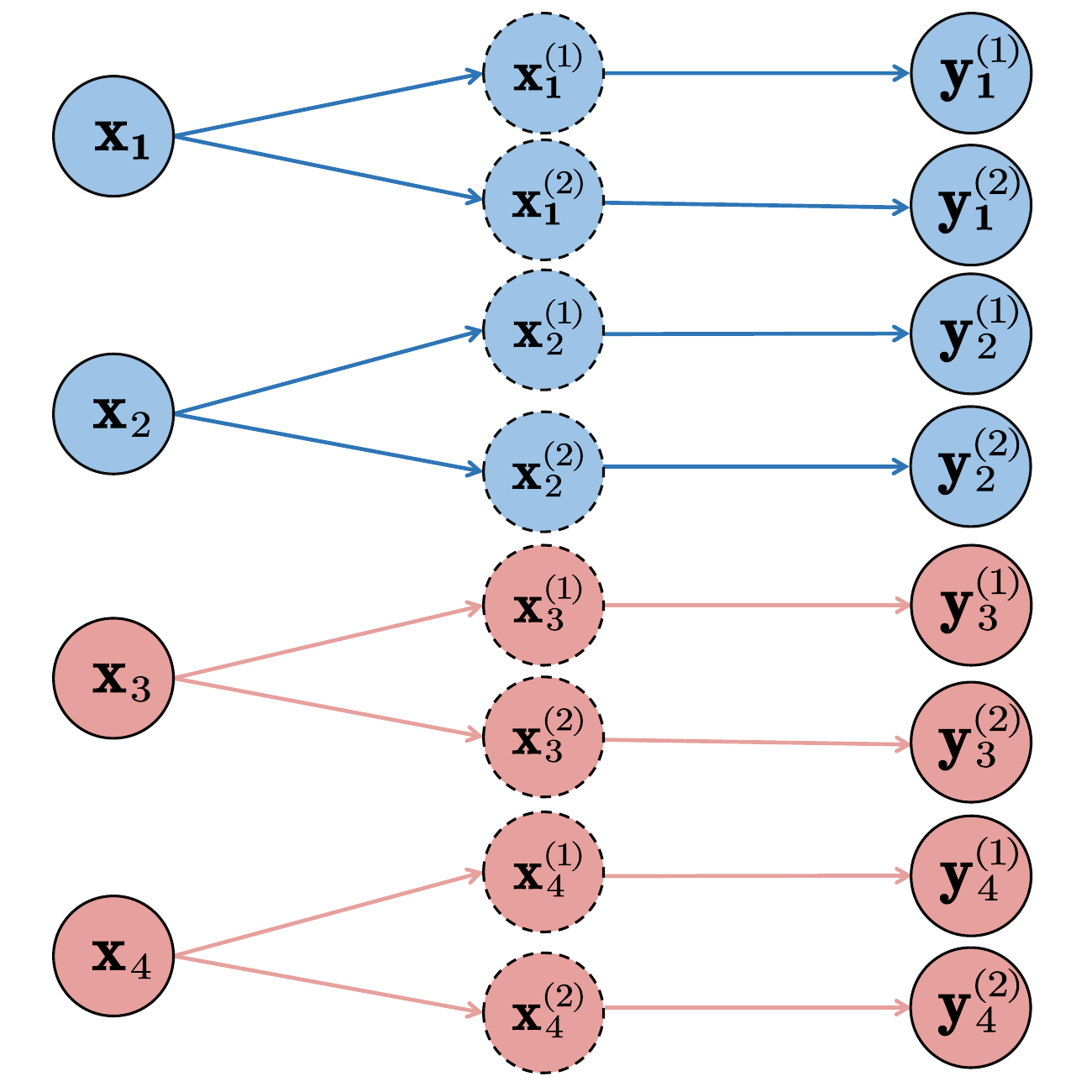}
        \caption{\SB}
        \label{fig:sb}
    \end{subfigure}
    \hfill %
    \begin{subfigure}{.32\textwidth}
        \centering
        \includegraphics[width=1.0\linewidth]{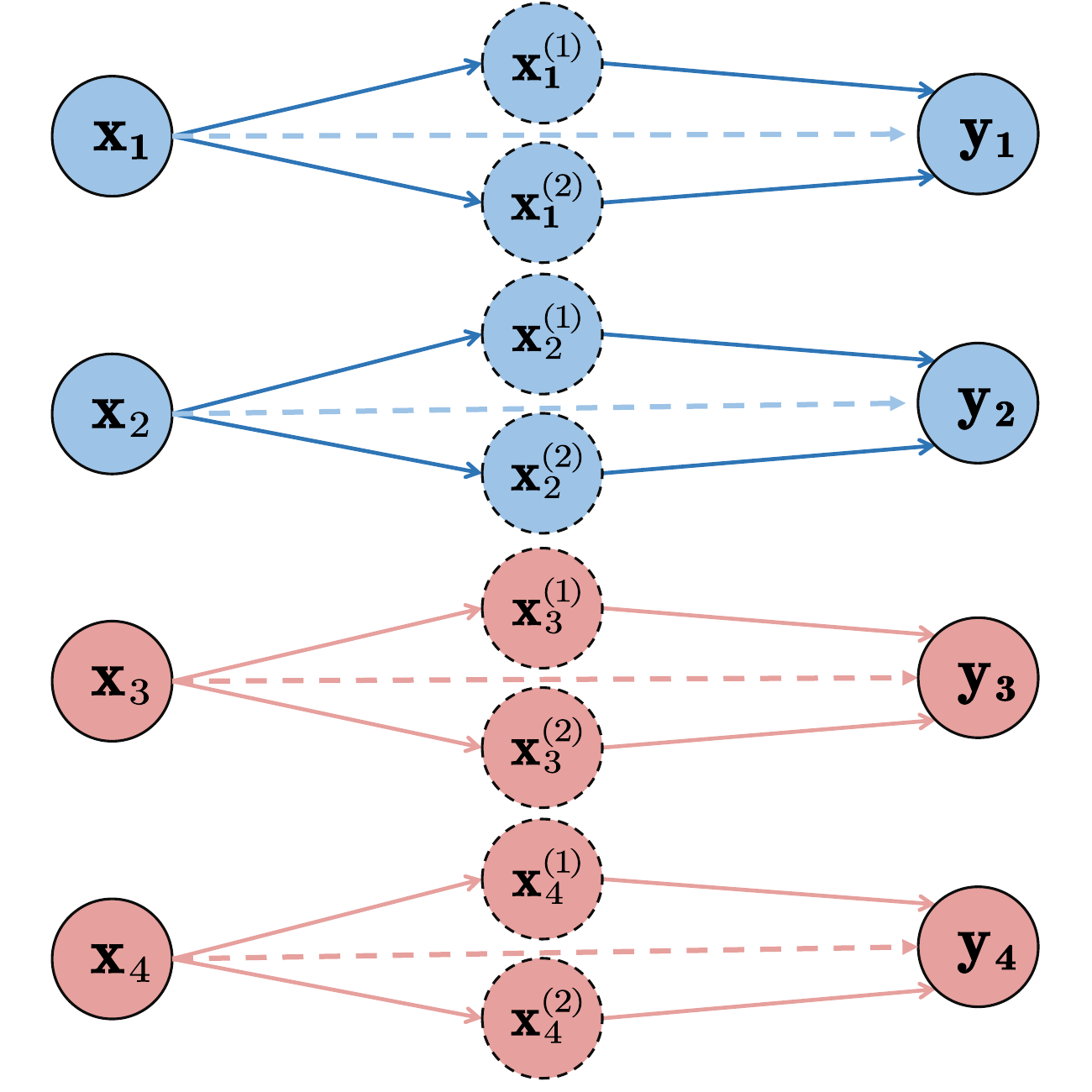}
        \caption{\SC}
        \label{fig:sc}
    \end{subfigure} %
    \caption{\textbf{Three different sample-to-target mapping strategies. } \textbf{\SA}: Multiple augmented samples within the same class are mapped to one same one-hot target.
        \textbf{\SB}: Each augmented sample is mapped to a unique soft target.
        \textbf{\SC}: Multiple augmented views of a sample are mapped to one same soft target.
        \textcolor{blue1}{Blue} and \textcolor{red1}{red} colors denote \textcolor{blue1}{class 1} and \textcolor{red1}{class 2}, respectively.
        For each strategy, $\xx_1$ and $\xx_2$ means original samples.
        The middle column illustrates the augmented samples produced by various augmentation strategies.
        On the left, $\yy_1$ and $\yy_2$ represent the targets associated with each augmented sample.
    }
    \vspace{-10pt}
    \label{fig:strategy}
\end{figure*}

\vspace{-10pt}
\section{Preliminary} \label{sec:definition}
\vspace{-10pt}
This section begins by formally defining three mapping strategies between samples $X$ and targets $Y$.
Subsequently, we critically analyze the limitations of existing loss functions in evaluating the impact of these mapping strategies on the representational capacity of models.
We then introduce a novel unified loss function.

\vspace{-10pt}
\subsection{Definition of Mapping Strategies} \label{sec:definition_strategy}
\vspace{-5pt}
Previous studies have demonstrated that data-efficient techniques, such as knowledge distillation and dataset distillation, which reconstruct the relationship between samples $X$ and targets $Y$, can significantly accelerate model training.
For instance, knowledge distillation leverages soft targets generated by teacher models to expedite the training process~\citep{yim2017gift}.
To thoroughly examine the relationship between samples $X$ and targets $Y$, we define mapping strategy, denoted as $\mpsi: \cX \rightarrow \cY$, where $\cX$ represents the data space and $\cY$ the target space.
Importantly, augmentations are crucial in deep learning, significantly boosting training effectiveness \citep{zhang2018mixup,geiping2022much}.
Consequently, this paper emphasizes the use of augmented samples.

\begin{figure*}[t]
    \centering
    \begin{subfigure}{.49\textwidth}  %
        \centering
        \includegraphics[width=1.0\linewidth]{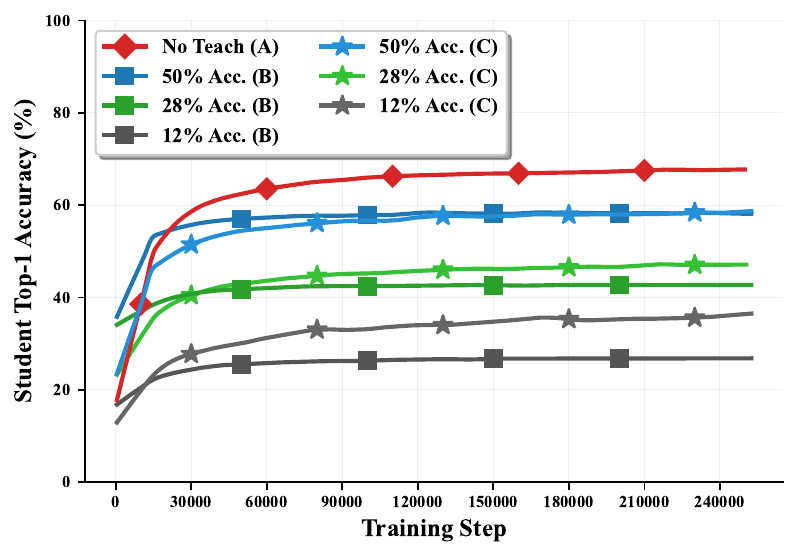}
        \captionsetup{labelformat=empty,skip=-8pt}
        \caption{}
        \vspace{-5pt}
    \end{subfigure}
    \hfill %
    \begin{subfigure}{.49\textwidth}
        \centering
        \includegraphics[width=1.0\linewidth]{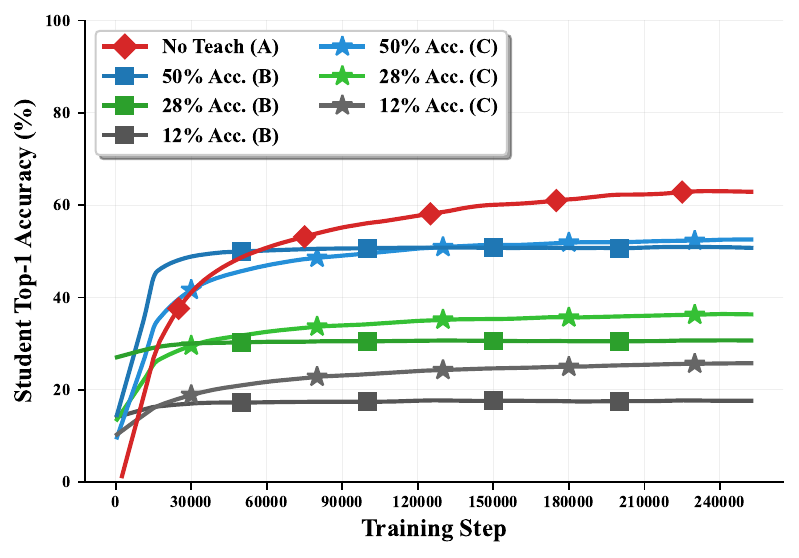}
        \captionsetup{labelformat=empty,skip=-8pt}
        \caption{}
        \vspace{-5pt}
    \end{subfigure}
    \caption{\textbf{Student training performance under three strategies using ResNet50 (Left) and ViT (Right) as backbone on ImageNet.} Under different backbones, \SA\ shows long-term advantages, while \SB\ exhibits short-term benefits. And the advantages of \SC\ become increasingly prominent when applied to weaker teacher models.}
    \vspace{-15pt}
    \label{fig:ImageNet}
\end{figure*}

\vspace{-10pt}
\paragraph{\SA: Mapping multiple augmented samples within the same class into a one-hot target.}
In conventional representation learning, data pairs $(X, Y)$ represent a non-injective mapping, denoted as \SA, as shown in \figref{fig:sa}.
For example, for a classification task, one class $c$ may include multiple samples $X_c$, indicating that multiple samples are mapped to a same class.
The mapping strategy is formalized as $\mpsi_\mathrm{A}(\xx_i^{(j)}) = c$, where $j$ denotes the $j$-th augmented sample of the original sample $\xx_i$.

\vspace{-10pt}
\paragraph{\SB: Mapping each augmented sample within the same class into a unique soft target.}
For efficient knowledge distillation and dataset distillation, these approaches reconstruct the sample-to-target relationship using more informative soft targets.
Each augmented sample corresponds to a unique soft target generated by a pre-trained model, establishing an injective mapping from augmented samples to soft labels, as shown in \figref{fig:sb}.
Specifically, the mapping strategy is formalized as:
$\mpsi_\mathrm{B}(\xx_i^{(j)}) = \yy_i^{(j)}$, where
$\yy_i^{(j)}$ represents the soft target generated by a pre-trained model for the augmented sample $\xx_i^{(j)}$.
It is obvious that \textit{\SB introduces soft targets, which includes more information, thereby accelerating the training process of the model.}

\vspace{-10pt}
\paragraph{\SC: Mapping multiple augmented views of one sample into a same soft target.}
While \SB provides valuable information through soft targets, it inevitably introduces noise by associating a single sample with multiple labels, potentially resulting in sub-optimal results.
To mitigate this issue while preserving the advantages of soft targets, we introduce a novel mapping strategy, \SC.
This approach associates all augmented samples of an original sample with a single soft target, as shown in \figref{fig:sc}.
It establishes a non-injective mapping from augmented samples to targets while maintaining an injective mapping from original samples to targets.

Specifically, the mapping function is defined as $\mpsi_\mathrm{C}(\xx_i^{(j)}) = \yy_i$, where $\yy_i$ is generated by a pre-trained model for the original sample $\xx_i$.
It preserves the rich information of soft targets while minimizing the noise introduced by multiple soft targets, which effectively balances the consistency of \SA with the variability of \SB.

\subsection{Design A Unified Loss Function}\label{sec:design_loss}
Having defined three distinct sample-to-target mapping strategies, we subsequently investigate their impact on the representational capacity of models.
Following previous benchmarks and research \citep{he2020momentum,chen2020simple, grill2020bootstrap}, the representation ability of a trained model
can be evaluated using an offline linear probing strategy, which involves decoupling the model into a backbone and a classifier.
However, two inherent challenges emerge when employing conventional loss functions:
(1) They inherently integrate these mapping strategies during training, complicating the evaluation of individual strategies.
(2) They obscure the backbone's representational ability due to classifier influence.
Therefore, in this subsection, we first review conventional loss functions and analysis and their limitations.
Subsequently, we propose a novel unified loss function to mitigate these challenges.

\begin{table}[!t]
    \centering
    \caption{\textbf{Comparison between the deconstructed loss function designed in our paper and traditional loss function.} The result shows the student performance using \SB under different teachers. Combining the backbone and classifier into a unified framework with a standard loss function will cause the accuracy of the student model consistently lower than the corresponding teacher.
    While with our proposed decoupled training framework, the student model can outperform the teacher, especially for weak teachers. \textit{(dataset: CIFAR-10, backbone: ResNet-18, training step: 50k)}}
    \vspace{5pt}
    \renewcommand{\arraystretch}{1.2}
    \setlength{\tabcolsep}{3pt} %
    \resizebox{\linewidth}{!}{ %
    \begin{tabular}{c|ccccccccc}
        \toprule
        \multirow{2}{*}{Loss} & \multicolumn{9}{c}{Teacher Top-1 Accuracy (\%) $\pm$1\%}                                                                                                                         \\
        \cmidrule(lr){2-10}
                                  & 10\%                       & 20\%                       & 30\%                       & 40\%                       & 50\%                       & 60\%                       & 70\%                       & 80\%                       & 90\%  \\
        \midrule
        Standard                  & 9.97 {\scriptsize$\pm$ 0.3} & 19.41 {\scriptsize$\pm$ 0.6} & 28.68 {\scriptsize$\pm$ 0.4} & 39.92 {\scriptsize$\pm$ 0.3} & 50.01 {\scriptsize$\pm$ 0.3} & 59.79 {\scriptsize$\pm$ 0.4} & 69.90 {\scriptsize$\pm$ 0.4} & 79.96 {\scriptsize$\pm$ 0.5} & 89.52 {\scriptsize$\pm$ 0.2} \\
        Ours                      & 35.96 {\scriptsize$\pm$ 0.4} & 39.70 {\scriptsize$\pm$ 0.4} & 41.61 {\scriptsize$\pm$ 0.5} & 46.11 {\scriptsize$\pm$ 0.3} & 57.18 {\scriptsize$\pm$ 0.4} & 63.90 {\scriptsize$\pm$ 0.5} & 72.58 {\scriptsize$\pm$ 0.2} & 82.93 {\scriptsize$\pm$ 0.4} & 90.35 {\scriptsize$\pm$ 0.3} \\
        Gain                      & \textbf{3.61}              & \textbf{2.04}              & \textbf{1.45}              & \textbf{1.16}              & \textbf{1.14}              & \textbf{1.07}              & \textbf{1.04}              & \textbf{1.04}              & \textbf{1.01}  \\
        \bottomrule
    \end{tabular}
    }
    \label{tab:loss}
    \vspace{-12pt}
\end{table}

\vspace{-6pt}
\subsubsection{Conventional Loss Functions}
\vspace{-4pt}
Cross-entropy (CE) loss is widely recognized as a standard objective for training deep neural networks on classification tasks.
However, CE loss continues to propagate one-hot target information to the model's backbone, implicitly introducing the multiple-to-one mapping to \SA.
This characteristic complicates the independent analysis of each mapping strategy with respect to representation ability, which is unsuitable for our objectives.
Furthermore, most knowledge distillation approaches \citep{gou2021knowledge,xu2020knowledge,zhao2022decoupled} treat the backbone and classifier as a unified entity during training.
This coupling potentially obscures the backbone's feature extraction capabilities, as they may be inevitability affected by the classifier's performance.
In summary, evaluating the impact of mapping strategies on representation ability of backbone remains challenging.

\vspace{-4pt}
\subsubsection{Separation Loss Function Design}
To address the challenges previously outlined, we propose a novel loss function designed to evaluate the influence of various strategies on representation ability.
Specifically, we employ the Kullback-Leibler (KL) Divergence to leverage the information embedded in soft targets during the training of the backbone.
For the classifier, we decouple it from the entire model and train it independently using cross-entropy (CE) loss.
This approach prevents the transmission of one-hot target information from the classifier to the backbone, thus maintaining the separation between these mapping strategies.
Furthermore, this decoupled training strategy ensures that the classifier performance does not adversely affect the representational capacity of backbone.
\looseness=-1

\vspace{-4pt}
\paragraph{Backbone Training.}
For backbone training, denoted as $\mphi_{\mtheta}$, we employ the Kullback-Leibler (KL) divergence to align its predictions with the soft targets generated by the teacher models, thus effectively incorporating the \SB and \SC schemes.

To decouple the effect of the classifier, we introduce a novel softmax layer $\gg$, projecting the outputs of $\mphi_{\mtheta}$ into soft targets.
This layer replaces the classifier, thereby ensuring the classifier's isolation from the training process.
The loss for backbone training is formulated as follows:
\begin{equation}
    \textstyle
    \cL_{\mathrm{B}}(\mphi_{\mtheta}, \gg) = \sum_{i=1}^{N} \sum_{j=1}^{n_i} D_{\mathrm{KL}} \left( \mpsi(\xx_i^{(j)}) \parallel \gg(\mphi_{\mtheta}(\xx_i^{(j)})) \right) \,,
\end{equation}
where $N$ is the number of data points, $n_i$ is the number of augmented samples for $\xx_i$, $\mpsi \in \{\SA, \SB, \SC\}$ represents the mapping strategy, and $\xx_i^{(j)}$ denotes the $j$-th augmented sample of the original sample $\xx_i$.

\paragraph{Classifier Training.}
Following previous works \citep{gou2021knowledge,xu2020knowledge}, we employ the Cross-Entropy (CE) loss to leverage one-hot targets in training the classifier $\hh$.
The loss for classifier training is defined as follows:
\begin{equation}
    \textstyle
    \cL_{\mathrm{H}}(\hh) = - \sum_{i=1}^{N}\sum_{j=1}^{n_i} y_i \log (\hh(\mphi_{\mtheta}(\xx_i^{(j)}))) \,,
\end{equation}
where $y_i$ is the one-hot target of the original sample $\xx_i$.

Finally, the total loss function for model training is the sum of the backbone and classifier losses:
\begin{equation}
    \textstyle
    \cL_{\mathrm{total}} = \cL_{\mathrm{B}}(\mphi_{\mtheta}, \gg) + \cL_{\mathrm{H}}(\hh) \,.
\end{equation}

\subsubsection{Loss Function Evaluation}
To further compare the differences between our designed loss function and traditional loss functions,
we first train a series of teacher models with varying performance (in terms of their validation accuracy).
Subsequently, these teacher models are employed to train student models to convergence, as shown in~\tabref{tab:loss}, 
Under the training framework of our decoupled loss function, the student models consistently outperform their corresponding teachers. 
And relatively weaker teacher models can significantly enhance the performance of student models.

\begin{figure}[t]
    \centering
    \adjustbox{valign=t}{%
      \begin{minipage}{0.45\textwidth} %
        \includegraphics[width=\linewidth]{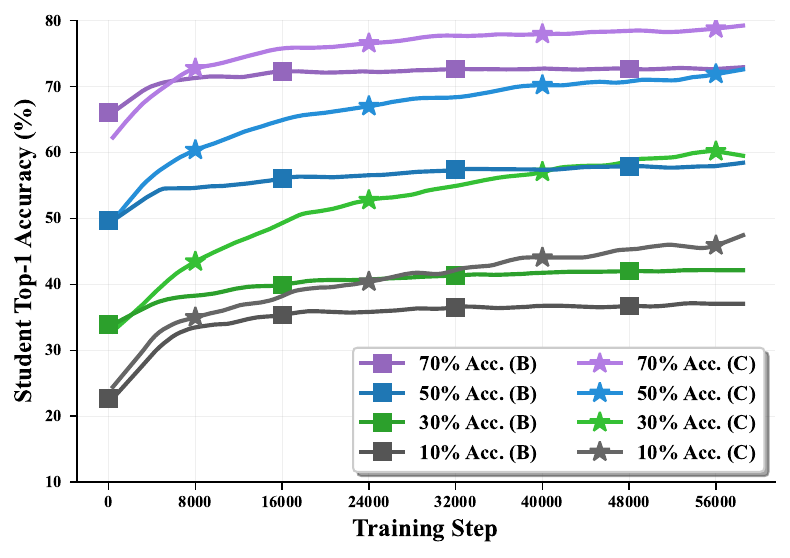}
      \end{minipage}%
    }
    \adjustbox{valign=t}{%
      \begin{minipage}{0.45\textwidth} %
        \centering
        \small
        \renewcommand{\arraystretch}{1.63} %
        \setlength{\tabcolsep}{3pt} %
        \resizebox{\linewidth}{!}{ %
        \begin{tabular}{cc|ccc}
          \toprule
          \multirow{2}{*}{Backbone} & \multirow{2}{*}{Strategy} & \multicolumn{3}{c}{Teacher Acc. (\%) $\pm$1\%} \\
          \cmidrule(lr){3-5}
          & & 20\% & 40\% & 60\% \\
          \midrule
          \multirow{3}{*}{MobileNetV2 \citep{sandler2018mobilenetv2}} 
          & B & 28.75 {\scriptsize$\pm$ 0.4} & 42.34 {\scriptsize$\pm$ 0.5} & 61.49 {\scriptsize$\pm$ 0.4} \\
          & C & 30.84 {\scriptsize$\pm$ 0.5} & 47.08 {\scriptsize$\pm$ 0.4} & 63.71 {\scriptsize$\pm$ 0.3} \\
          & Gain & \textbf{1.07} & \textbf{1.11} & \textbf{1.04} \\
          \midrule
          \multirow{3}{*}{EfficientNet \citep{tan2019efficientnet}} 
          & B & 23.24 {\scriptsize$\pm$ 0.3} & 42.12 {\scriptsize$\pm$ 0.4} & 61.47 {\scriptsize$\pm$ 0.5} \\
          & C & 25.01 {\scriptsize$\pm$ 0.6} & 47.20 {\scriptsize$\pm$ 0.4} & 63.08 {\scriptsize$\pm$ 0.3} \\
          & Gain & \textbf{1.08} & \textbf{1.12} & \textbf{1.03} \\
          \bottomrule
        \end{tabular}
        }
      \end{minipage}%
    }
    \caption{\textbf{Comparison of \SB and \SC.} 
    \textbf{Left:} \SC demonstrates superior performance in long-term training. Even when using a randomly initialized teacher model with 10\% accuracy, the student model trained with \SC outperforms the student model trained under \SB with a teacher model of 30\% accuracy. 
    \textbf{Right:} Results are consistent on different backbones. \textit{(training step: 39k)}}
    \label{fig:SBC}
    \vspace{-12pt}
\end{figure}

\section{What role do targets play? A Comprehensive Investigation of Mapping Strategies}

This section systematically investigates the impact of various mapping strategies $\mpsi : \cX \rightarrow \cY$ and different target generation on the representational capacity of the backbone network $\mphi_{\mtheta}$.
For experimental setup, see \appref{app:details}.
We begin by comparing and analyzing how different target types, namely one-hot targets in \SA, soft targets in \SB, and our proposed \SC, affect the representational capacity of $\mphi_{\mtheta}$, as detailed in \secref{sec:exp1}.
Furthermore, we assess the effect of soft targets generated by teachers with different performance, as discussed in \secref{sec:exp2}.
Finally, we explore the impact of soft targets produced by teacher models trained with diverse augmentation strategies, as described in \secref{sec:exp3}.

\vspace{-2pt}
\subsection{Comparing analysis of Target mapping: What kinds of Targets Drives Better Performance?} \label{sec:exp1}

\begin{figure*}[!t]
    \centering
    \begin{subfigure}{.49\textwidth}  %
        \centering
        \includegraphics[width=1.0\linewidth]{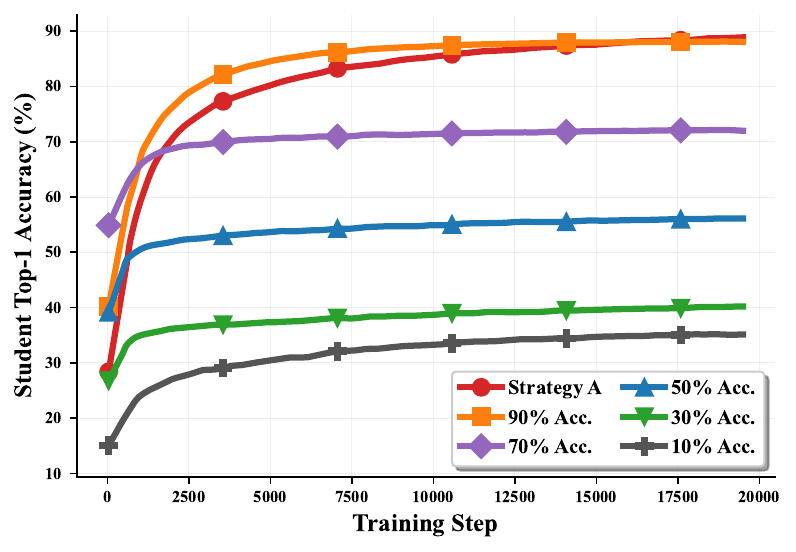}
        \caption{Comparison \SA and \SB.}
        \vspace{-5pt}
        \label{fig:SAB}
    \end{subfigure}
    \hfill %
    \begin{subfigure}{.49\textwidth}
        \centering
        \includegraphics[width=1.0\linewidth]{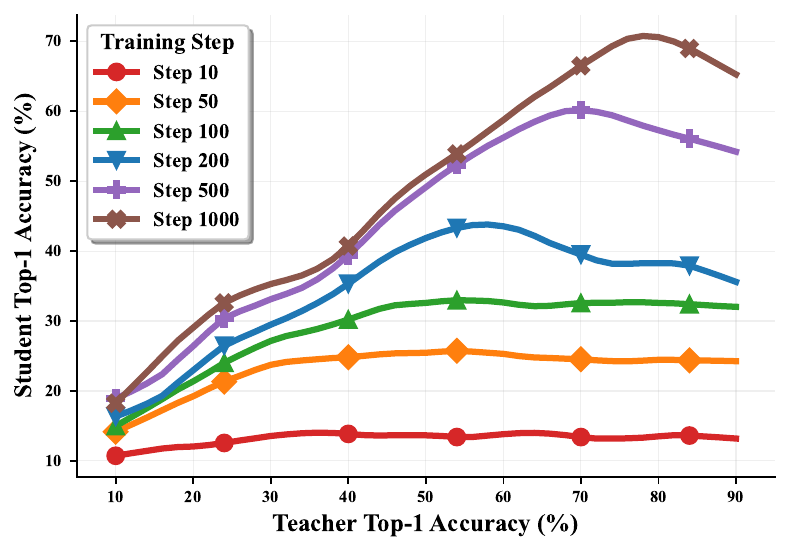}
        \caption{Early-stage Training in \SB.}
        \vspace{-5pt}
        \label{fig:task2}
    \end{subfigure}
    \caption{\textbf{\SB accelerates early-stage training but achieves inferior final performance.} \textbf{Left:} Comparison of \SA and \SB that employs teacher models of differing accuracies. %
        \textbf{Right:} Weaker teacher models can facilitate early-stage learning. %
        \textit{(dataset: CIFAR-10, backbone: ResNet-18.)}}
        \label{fig:task12}
    \vspace{-12pt}
\end{figure*}

\paragraph{\SB: more targets, faster convergence.}
We first use a series of teacher models with various accuracices to train the student.
The results, as presented in~\figref{fig:SAB}, demonstrate that \textit{models trained with \SB exhibit accelerated learning during early training stages.}
This is due to the additional information provided by soft targets. %
Conversely, models trained with the traditional \SA, despite slower improvements in the early-stage training, continue to improve performance over time and ultimately surpass the models trained with \SB.

\paragraph{\SC: Fewer Targets, higher performance.}

Though both \SB and \SC using soft labels generated by a pre-trained teacher model.
The key distinction lies in the number of targets used during model training: \SB assigns different soft targets for each augmented sample, whereas \SC utilizes the same soft target for all augmented samples derived from the same original sample.
We conduct extensive experiments to assess these strategies with different teachers on different datasets and backbones.
As shown in~\figref{fig:ImageNet},~\figref{fig:SBC} and numerical results in~\tabref{tab:ImageNet},~\tabref{tab:ImageNet_AC},
The results show that though in the early training phase, \SB exhibits an advantage when the teacher model's accuracy is relatively high.
As training progresses, the student model trained using \SC generally achieves higher gains.
And compared to traditional \SA, \SC can also retain the ability to accelerate early stage training.
For CIFAR10, We further conduct experiments employing a teacher model with different accuracy and compare the performance of the student model at different training stages under these two strategies, as illustrated in \figref{fig:CIFAR10_BC_compare} and~\tabref{tab:CIFAR10_compareBandC}. %
We find that as the teacher model's accuracy increases from 10\% to 90\%, the improvement in student model performance initially rises and then declines.
Notably, with a teacher model accuracy of 30\%, the performance gain (a factor of 1.35 increase) achieved with \SC is substantially greater than that with a 90\% accuracy teacher model (a factor of 1.01 increase).
These results suggest that the benefit of \SC diminishes when the teacher model's accuracy is sufficiently high.

\vspace{10pt}
\begin{claim}
    \SB accelerates the early-stages training but ultimately limits accuracy in relation to teacher accuracy.
    Compared to \SB, \SC consistently achieves higher final converged accuracy while maintaining short-stage acceleration. %
\end{claim}

\vspace{-5pt}
\subsection{Early-Stage Learning: Stronger Teachers Are Not Always Optimal.} \label{sec:exp2}
Previous experiments have demonstrated that the performance of student models is typically influenced by the targets generated by different teacher models, with strong teachers generally improving the final accuracy of the student model.
However, it is crucial to note that \textit{stronger teacher models are not universally advantageous.}
In this experiment, we focus on the early stages of student training.
We fix the number of training steps\footnote{We also experimented with more training steps, see \tabref{tab:CIFAR10_diff_step} in \appref{app:results} for details.} to examine whether stronger teachers lead to greater improvements.
As illustrated in \figref{fig:task2}, for a fixed number of training steps, as the teacher model's accuracy increases, there is an initial improvement followed by a decline in the performance of student models. This suggests that strong teacher models do not consistently provide significant advantages and may even hinder student model training during early-stage learning.

\vspace{5pt}
\begin{claim}
    While stronger teacher models generally enhance student model performance over the long term, weaker teacher models can facilitate early-stage learning, particularly when the number of training steps is limited.
\end{claim}

\begin{figure*}[!t]
    \centering
    \begin{subfigure}{.49\textwidth}
        \centering
        \includegraphics[width=1.0\linewidth]{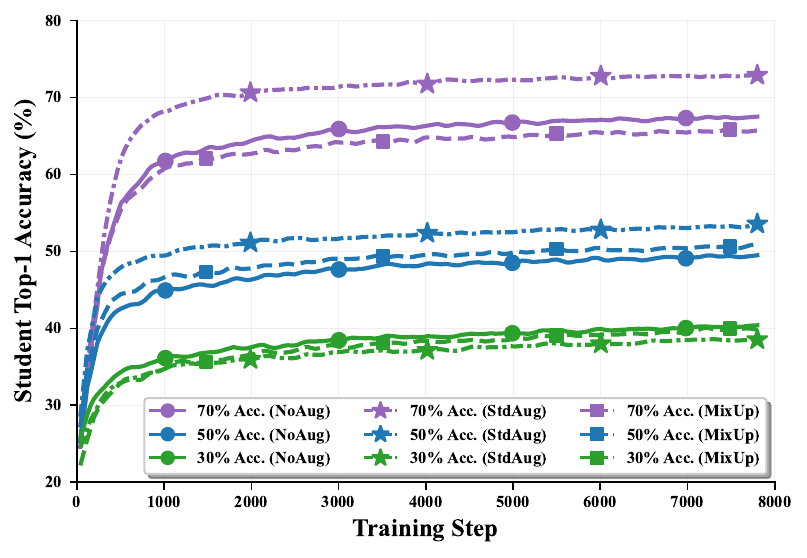}
        \caption{Data Augmentations for Teacher Model}
        \vspace{-5pt}
        \label{fig:task31}
    \end{subfigure}
    \hfill
    \begin{subfigure}{.49\textwidth}
        \centering
        \includegraphics[width=1.0\linewidth]{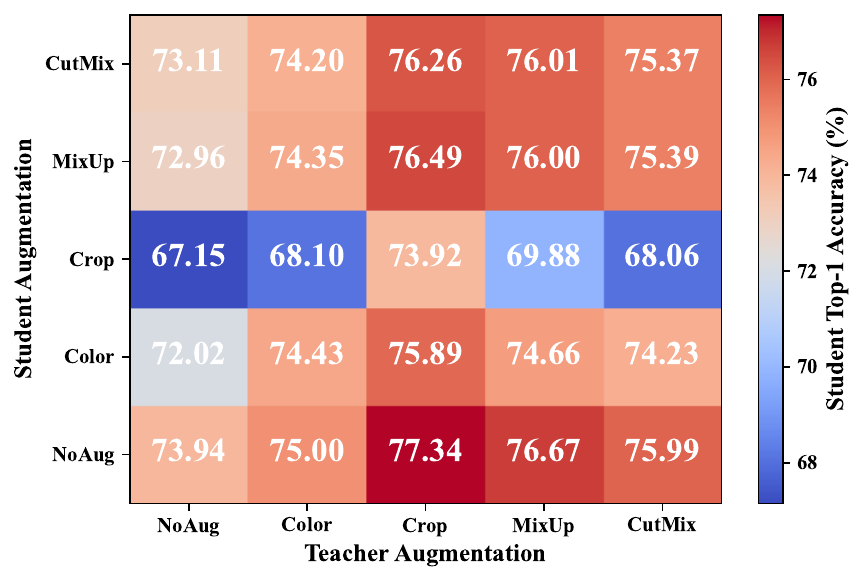}
        \caption{Data Augmentations for Teacher and Student Model}
        \vspace{-5pt}
        \label{fig:task32}
    \end{subfigure}
    \caption{\textbf{Left: Teacher models trained with various augmentation strategies.}
        Teacher models with high accuracy, trained using MixUp, can not help student model training.
        \textbf{Right: Applying various augmentations to both teacher and student models.} RandomResizedCrop significantly reduces student model performance.}
    \vspace{-15pt}
    \label{fig:exp_aug}
\end{figure*}

\vspace{-5pt}
\subsection{Impact of Data Augmentation on Soft Target Generation} \label{sec:exp3}

Prior sections have concentrated on the influence of teacher model performance on student model training.
In this section, we focus on the intrinsic properties of teacher models.
A notable study by \citep{zhang2018mixup} highlights that advanced augmentation techniques like MixUp can enhance a model's generalization and robustness by creating new training samples through linear interpolation between pairs of examples and their corresponding labels.
Therefore, this section explores the influence of data augmentation techniques employed during teacher model training.

\vspace{-5pt}
\paragraph{Teacher Models with Various Data Augmentations.}
In this experiment, we use teacher models with three identical performance levels ($30$\%, $50$\%, and $70$\%) but trained with different augmentation strategies:
(i) \textbf{NoAug:} no augmentation, (ii) \textbf{StdAug:} standard augmentation without MixUp, and (iii) \textbf{MixUp:} MixUp augmentation.
Detailed descriptions of these strategies are provided in \appref{app:details}.
Additional augmentation strategies applied to the teacher model are discussed in \tabref{tab:CIFAR10_diff_aug} in \appref{app:results}.

The results, as depicted in \figref{fig:task31}, demonstrate that \textit{for high-accuracy teacher models, applying MixUp augmentation to the teacher does not benefit the training of the student model.}
Interestingly, when teacher model accuracy is relatively low (e.g., 30\% accuracy, as shown by the \textcolor{green1}{green line}), the advantage of standard and MixUp strategies diminishes, and models trained with no augmentation show the best performance.

\vspace{-5pt}
\paragraph{Both student and Teacher Models with Various Data Augmentations.}
To further assess the impact of augmentation techniques applied on student models themself, we fix the teacher model's accuracy\footnote{Results for teacher models with different accuracy levels are available in \figref{fig:CIFAR10_heatmap_appendix} in \appref{app:results}.} at $70$\% and train the student model using various data augmentations.
The data augmentation techniques include five strategies: NoAug, RandomResizedCrop\footnote{RandomResizedCrop is notably regarded as a strong augmentation method, alongside mix-based strategies.}, StdAug, MixUp, and CutMix~\citep{yun2019cutmix}.
Detailed descriptions of these augmentation strategies are provided in \appref{app:details}.
The results, shown in \figref{fig:task32}, reveal that:
\vspace{-5pt}
\begin{enumerate}[label=(\alph*), leftmargin=16pt]
    \item For the teacher model, employing RandomResizedCrop, MixUp, or CutMix significantly enhances student model accuracy compared to other methods.
    \item For the student model, using RandomResizedCrop significantly reduces performance unless the teacher model was also trained with RandomResizedCrop.
\end{enumerate}

\begin{claim}
    (1) Teacher models trained with augmentation strategies such as MixUp can not consistently enhance student model performance, and no augmentation for the weaker teacher better facilitates student training.. %
    (2) Additionally, employing RandomResizedCrop for student model training significantly impairs performance unless consistently applied to both teacher and student model training.
\end{claim}

\vspace{-5pt}
\section{How Does Sample Variation Affect Performance?} \label{sec:sample}
\vspace{-5pt}

In preceding sections, we focus on how variations in target mappings impact the model training process.
Notably, equally crucial to the performance of deep learning models is the quantity and quality of training samples.
In this section, we first investigate how sample quantity affects model performance, followed by an analysis of sample quality.
\footnote{The accuracy of the teacher models in \secref{sec:ipc} and \secref{sec:student_aug} is 80\% and 90\%, respectively. Additional experiments with different teacher model accuracies are detailed in~\figref{fig:CIFAR10_BC_diff_teacher} and~\figref{fig:CIFAR10_B_student_aug} in \appref{app:results}.}.

\vspace{-10pt}

\subsection{Less is More? Probing Data Quantity through IPC Variations} \label{sec:ipc}

\begin{figure*}[!t]
    \centering
    \begin{subfigure}{.49\textwidth}  %
        \centering
        \includegraphics[width=1.0\linewidth]{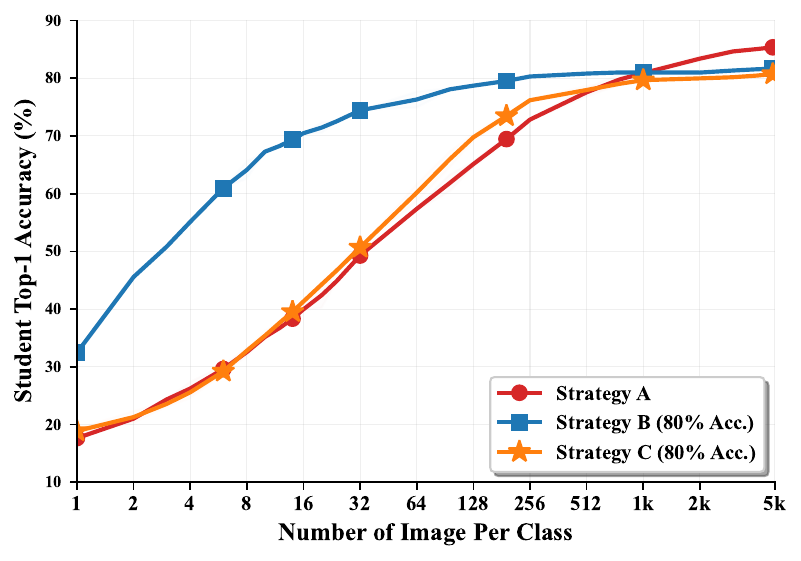}
        \caption{Influence of \IPC}
        \vspace{-5pt}
        \label{fig:task41}
    \end{subfigure}
    \hfill
    \begin{subfigure}{.49\textwidth}
        \centering
        \includegraphics[width=1.0\linewidth]{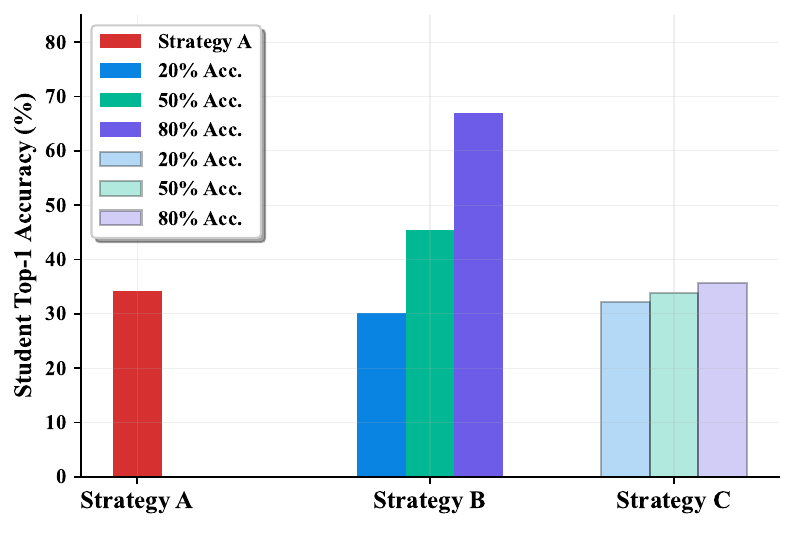}
        \caption{Strategies Comparison with extremely small \IPC}
        \vspace{-5pt}
        \label{fig:task42}
    \end{subfigure}
    \caption{\textbf{Left: Student model accuracy scaling with increasing Images Per Class (\IPC).}
        With limited IPC, \SB deviates from the typical power-law scaling compared with other strategies, whereas \SA exhibits notable advantages when \IPC is sufficient. \textbf{Right: Strategy comparison when \IPC = 10.} \SB demonstrates significant benefits, while \SC shows limited advantages.}
    \label{fig:exp_ipc}
    \vspace{-15pt}
\end{figure*}

\vspace{-5pt}
\paragraph{Influence of \IPC on Three Mapping Strategies.}
We begin by investigating how sample quantity influences model performance across three strategies using Images Per Class (\IPC) as a metric.
For each \IPC configuration, student models are trained using these three strategies for $10{,}000$ steps, a point at which models nearly achieve convergence.
The results are depicted in \figref{fig:task41}.
Key observations include:
\vspace{-5pt}
\begin{enumerate}[label=(\alph*), leftmargin=16pt]
    \item Both \SA and \SC exhibit typical \textit{power-law scaling}.
          As the number of images per class increases, the rate of improvement significantly diminishes.
          Notably, \SC demonstrates a slight advantage over \SA when the sample size is limited, such as when \IPC is below $500$.
    \item For \SB, \textit{limited \IPC results in substantial performance gains}, achieving approximately 30\% higher margins than \SA and \SC.
          However, as \IPC increases, the benefits diminish.
          For example, when \IPC surpasses 1K, the student model's accuracy approaches that of the teacher model and is eventually surpassed by \SA.
\end{enumerate}

\vspace{-15pt}
\paragraph{Extremely small \IPC Across Three Strategies.}
We further conduct experiments with the three strategies under extremely limited samples (e.g., \IPC = 10).
For \SB and \SC, we employ teacher models with three different performance levels (20\%, 50\%, and 80\%), all trained with identical augmentation.
As illustrated in \figref{fig:task42}, the results indicate that (1) with \SB, the student model nearly achieves the teacher model's accuracy, and teacher models with higher accuracy provide a greater advantage.
(2) Conversely, with \SC, the student model performance exhibits negligible differences despite the varying accuracies of the teacher models.
\looseness=-1

\vspace{5pt}
\begin{claim}
    (1) Student models trained with \SB in low \IPC achieve significantly higher accuracy compared to \SA and \SC.
    Moreover, teacher models with higher accuracy provide a greater advantage in \SB.
    (2) Yet, as sample size increases, the traditional \SA demonstrates superior performance, as both \SB and \SC become limited by the teacher model's accuracy.
\end{claim}

\vspace{-5pt}
\subsection{Strategic Enhancements: Data Augmentation for Superior Training} \label{sec:student_aug}

\begin{figure*}[t]
    \centering
    \begin{subfigure}{.32\textwidth}  %
        \centering
        \includegraphics[width=1.0\linewidth]{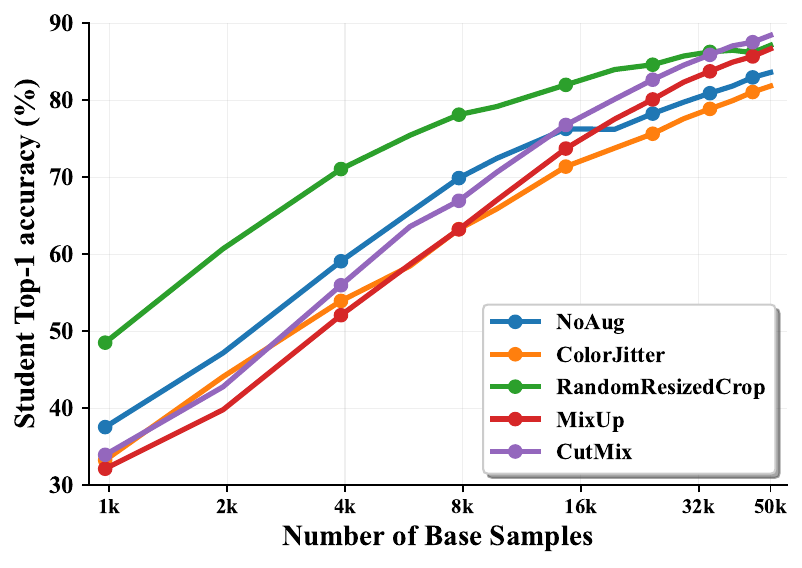}
        \caption{\SA}
        \vspace{-5pt}
        \label{fig:task51}
    \end{subfigure}
    \hfill %
    \begin{subfigure}{.32\textwidth}
        \centering
        \includegraphics[width=1.0\linewidth]{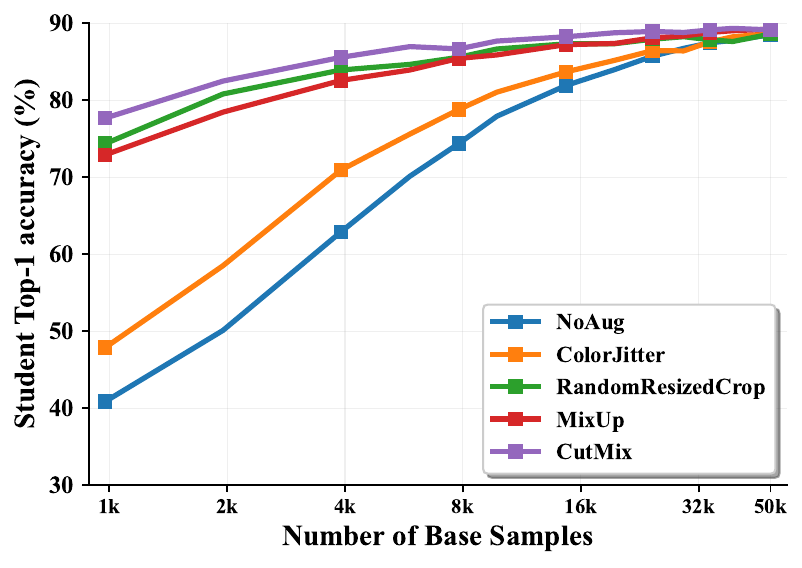}
        \caption{\SB}
        \vspace{-5pt}
        \label{fig:task52}
    \end{subfigure}
    \hfill %
    \begin{subfigure}{.32\textwidth}
        \centering
        \includegraphics[width=1.0\linewidth]{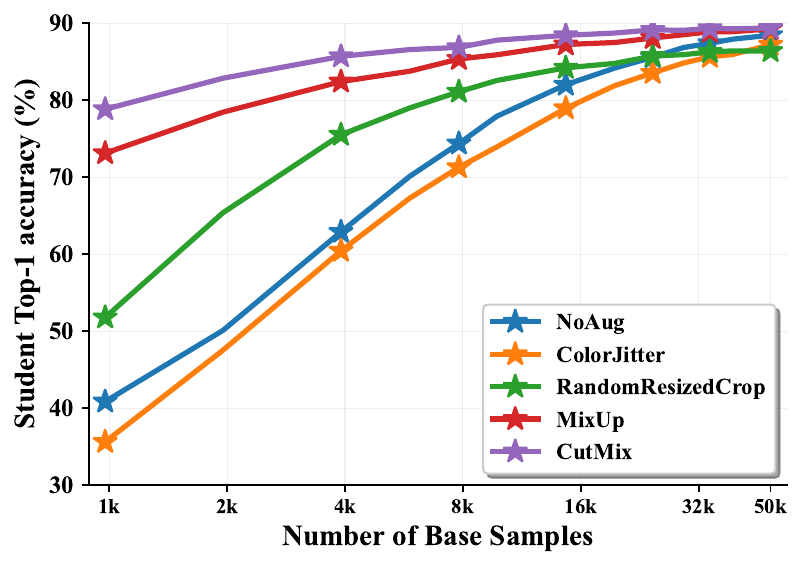}
        \caption{\SC}
        \vspace{-5pt}
        \label{fig:task53}
    \end{subfigure}
    \caption{
        \textbf{Scaling behavior of models using various data augmentation methods across three strategies.}
        For \SA, RandomResizedCrop enhances model performance significantly.
        In \SB and \SC, augmentation strategies surpass scaling law limitations when samples are limited.
        However, with sufficient samples, models converge to similar accuracy as the teacher model under both strategies, regardless of the augmentation used.
    }
    \vspace{-10pt}
    \label{fig:task5}
\end{figure*}

Beyond sample quantity, sample quality substantially impacts model performance.
Data augmentation techniques~\citep{perez2017effectiveness,shorten2019survey}, such as random cropping, rotation, flipping, and color jittering, introduce controlled transformations that increase the diversity of training data.
\citep{geiping2022much} investigates the relationship between data augmentations and scaling laws and quantify the value of augmentations.
And in this section, we evaluates the impact of various data augmentation strategies%
For each strategy, we vary the number of samples per class (\IPC) and apply different data augmentation methods to assess their influence on model training.
The results, depicted in \figref{fig:task5}, are as follows:
\vspace{-8pt}
\begin{itemize}[left=0.5cm, leftmargin=16pt]
    \item \SA (\figref{fig:task51}): The influence of augmentation strategies varies with sample size.
          RandomResizedCrop significantly outperforms other augmentation strategies (ColorJitter, MixUp, CutMix) regardless of sample size.
          Notably, when the sample size is limited, no augmentation except RandomResizedCrop proves beneficial.
          However, when the sample size is sufficient, e.g., \IPC exceeds 16K, CutMix and MixUp provide slight advantages over no augmentation.
    \item \SB (\figref{fig:task52}): All data augmentation strategies enhance model performance regardless of sample size.
          In particular, MixUp, CutMix, and RandomResizedCrop provide significant improvements with limited samples.
          Interestingly, with abundant samples, augmentation strategies converge to similar performance levels as that of the teacher model.
    \item \SC (\figref{fig:task53}):
          No Augmentation, ColorJitter, and RandomResizedCrop exhibit a performance pattern akin to \SA, while MixUp and CutMix show scaling behavior similar to \SB.
\end{itemize}
\vspace{5pt}
\begin{claim}
    (1) For \SA, RandomResizedCrop is the most effective augmentation method, particularly with limited sample sizes, whereas other strategies may not be beneficial.
    (2) Both \SB and \SC can surpass traditional \SA scaling law limitations through augmentation when data is scarce.
    With sufficient data, models utilizing \SB converge to similar performance levels regardless of the augmentation method employed.
\end{claim}

\section{Conclusion and Future Work} \label{sec:future}
\paragraph{Conclusion.}
To the best of our knowledge, this study is \uline{the first to emphasize the critical role of samples, targets, and their mappings in efficient learning}.
We present three sample-to-target mapping strategies and propose a unified framework for evaluation. %
Our analysis systematically investigates the impact of target and sample types, along with their quantities and qualities on model training.
Based on this analysis, we provide six key insights for improving training efficiency.

\paragraph{Future Work.}
The experiments presented in this paper are conducted \textit{within the context of computer vision}.
In future, we aim to perform a more comprehensive analysis of these phenomena \textit{theoretically} and \textit{experiment on other modality}, such as text data. %
By examining results across diverse datasets, networks and advanced algorithms, we will assess the robustness and generalizability of our findings.

\clearpage

%% file: resources/appendix.tex
\newpage

\section{Related Work} \label{app:related}
To assess the impact of data---consisting of samples $X$ and targets $Y$---on the learning, we revisit three distinct paradigms in deep learning and examine their respective data structures.

\vspace{-5pt}
\paragraph{Conventional representation learning.}
Representation learning~\citep{zhang2018network} enables models to automatically discover and extract meaningful features from raw data, facilitating tasks such as classification~\citep{li2022survey}, clustering~\citep{ren2024deep}, and segmentation~\citep{minaee2021image}.
These tasks typically rely on datasets with explicit sample-target pairs, such as ImageNet-1K~\citep{deng2009imagenet}.
Recent studies~\citep{radford2021learning,hafner2021clip} have extended this framework by utilizing text-image pairs to train robust representation models.
In these approaches, text is implicitly used as a target through contrastive learning~\citep{chen2020simple,khosla2020supervised}.
Moreover, self-supervised learning methods~\citep{grill2020bootstrap,zbontar2021barlow,chen2021empirical,caron2021emerging} have been developed to extract representations from unlabeled data by generating supervision signals through mechanisms like momentum models.

\textbf{Summary.} \textit{Current representation learning techniques construct learning targets \(Y\) from samples \(X\), explicitly or implicitly, to guide models in learning the nonlinear mapping between them.}

\vspace{-5pt}
\paragraph{Soft label learning via knowledge distillation.}
Knowledge distillation, introduced by~\citep{hinton2015distilling}, is a prominent technique for model compression in which a smaller student model learns from the outputs, or soft labels, of a larger teacher model.
This approach effectively transfers knowledge with minimal performance degradation, enabling smaller, efficient models while maintaining capability~\citep{gou2021knowledge}.
Subsequent research has expanded its application to enhance model robustness~\citep{xu2020knowledge}, facilitate domain adaptation~\citep{he2020knowledge}, and support semi-supervised learning~\citep{chen2020semi}.
Moreover, \citep{yim2017gift} demonstrates that knowledge distillation can accelerate the optimization process during training.

\textbf{Summary.} \textit{Knowledge distillation generates unique \textbf{soft targets} \(Y\) for each sample via a pre-trained teacher model, embedding rich information that enables the student model to closely mimic the behavior of the teacher.}

\vspace{-5pt}
\paragraph{Data-efficient learning through condensed data.}
Traditional dataset distillation aims to replicate the original dataset's behavior within a distilled dataset by minimizing discrepancies between surrogate neural network models trained on both synthetic and original data.
Key metrics in this process include matching gradients~\citep{zhao2020dataset}, features~\citep{wang2022cafe}, distributions~\citep{zhao2023dataset}, and training trajectories~\citep{cui2022dc}.
However, these methods incur substantial computational overhead due to the ongoing calculation of discrepancies, requiring numerous iterations for optimization and convergence.
This makes them challenging to scale to large datasets, such as ImageNet \citep{deng2009imagenet}.
An effective strategy involves identifying metrics that capture critical dataset information \citep{yin2023squeeze,sun2024diversity}, thereby eliminating the need for exhaustive comparisons between original and distilled datasets. Consequently, these methods scale efficiently to large datasets, such as ImageNet-1K.
For instance, SRe$^2$L~\citep{yin2023squeeze} compacts the entire dataset into a model, such as pre-trained neural networks like ResNet-18~\citep{he2016deep}, and subsequently extracts knowledge into images and targets, forming a distilled dataset.
Recently, RDED~\citep{sun2024diversity} suggests that images accurately recognized by strong observers, such as humans and pre-trained models, are pivotal for effective learning.

\textbf{Summary.} \textit{Dataset distillation methods extract key information from the original dataset, forming more informative samples \(X\) and targets \(Y\) that facilitate data-efficient learning.}

\section{Experimental Setup} \label{app:details}

\paragraph{Dataset and Network.}
We conduct experiments using both large-scale and small-scale datasets, specifically CIFAR-10 \citep{krizhevsky2009learning}, CIFAR-100 \citep{krizhevsky2009learning}, Tiny-ImageNet \citep{le2015tiny} and ImageNet-1k \citep{imagenet_cvpr2009}.
In the main body, our model utilizes ResNet-18 \citep{he2016deep} as the backbone, selected due to its widespread use and its generalizable performance across different architectures \citep{huang2017densely}.
We also evaluate MobileNetV2 \citep{sandler2018mobilenetv2}, EfficientNet \citep{tan2019efficientnet}, ResNet-50 and Vision Transformer \citep{dosovitskiy2021image} as the backbone on the proposed \SC.
The evaluation metric for all experiments is the Top-1 classification accuracy (\%) on the test set.
In the main body, we primarily present results on CIFAR-10 and ResNet-18, while others are provided in \appref{app:diff_datasets}.

\begin{figure}[h!]
    \centering
    \begin{subfigure}{.49\textwidth}  %
        \centering
        \includegraphics[width=1.0\linewidth]{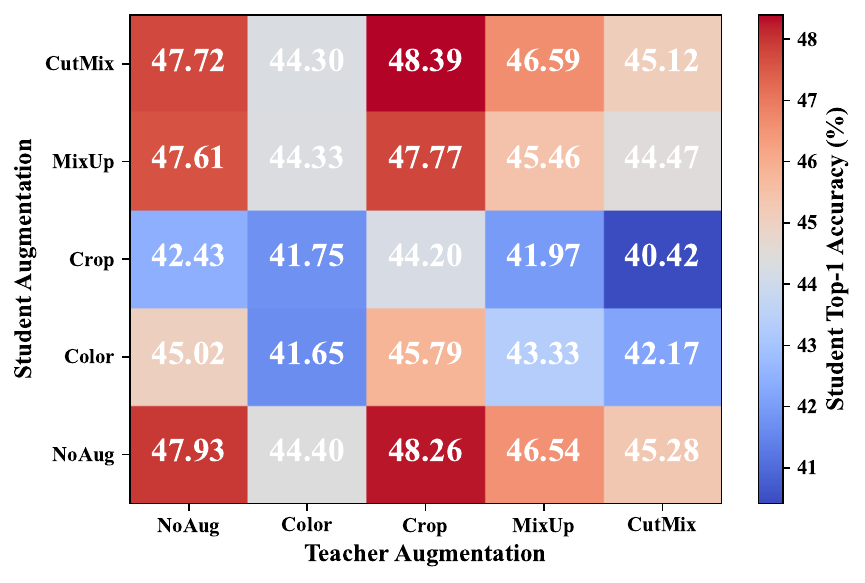}
        \captionsetup{labelformat=empty,skip=-8pt}
        \caption{}
        \label{fig:CIFAR10_heatmap30}
    \end{subfigure}
    \hfill %
    \begin{subfigure}{.49\textwidth}
        \centering
        \includegraphics[width=1.0\linewidth]{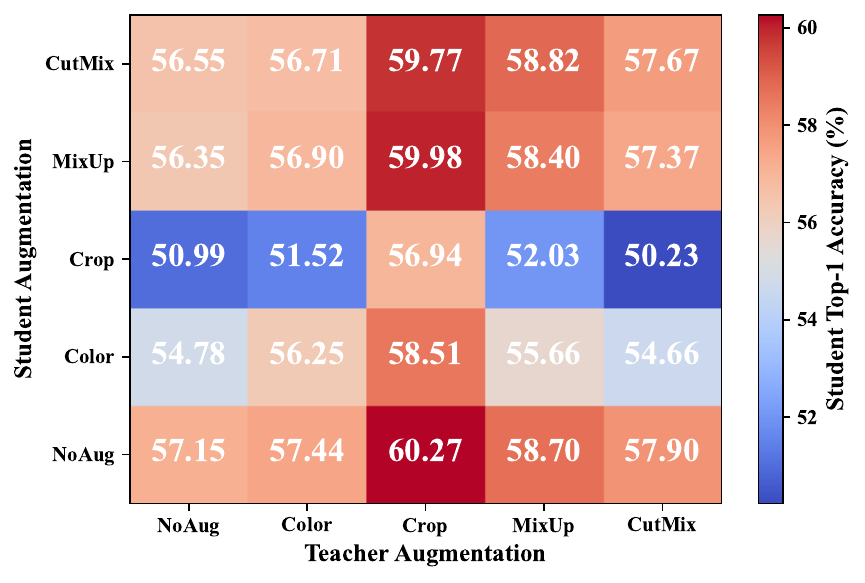}
        \captionsetup{labelformat=empty,skip=-8pt}
        \caption{}
        \label{fig:CIFAR10_heatmap50}
    \end{subfigure}
    \caption{\textbf{Ablation experiments of training the student model with 30\% (\textbf{Left}) and 50\% (\textbf{Right}) teacher models in the main body.} It is evident that using the RandomResizedCrop strategy has certain disadvantages for the student model. Additionally, applying the MixUp strategy to teacher models with lower accuracy undermines the effectiveness of training the student model compared to NoAug.}
    \label{fig:CIFAR10_heatmap_appendix}
\end{figure}

\paragraph{Batch size and training step settings.}
All experiments in the main body were conducted using the CIFAR-10 dataset.
By default, the number of images per class in the training set is 5000, except for experiments in~\secref{sec:ipc},~\secref{sec:student_aug}.
For all experiments throughout the paper, samples were randomly shuffled at the beginning of each epoch, and the batch size was fixed at 128.
The number of batches per epoch after applying drop last strategy is 390 for both CIFAR10 and CIFAR100, and 780 for TinyImageNet.
For the experiments in Figure~\ref{fig:task12} and~\ref{fig:dataset_exp1},
training was run for 50 epochs (corresponding to 20,000 steps for CIFAR10 and CIFAR100, and 40,000 steps for TinyImageNet), with validation averaged 10 times per epoch.
For the experiments in Figure~\ref{fig:task31} and~\ref{fig:dataset_teacher_diff_aug}, training was run for 20 epochs (7,800 steps for CIFAR10 and CIFAR100, and 15,600 steps for TinyImageNet), also with validation averaged 10 times per epoch.
For the experiments in Figure~\ref{fig:CIFAR10_BC_compare}, training was run for 150 epochs (i.e., 58,500 steps), with validation performed once at the end of each epoch (i.e., every 390 steps).
And for experiments in~\secref{sec:ipc},~\secref{sec:student_aug} and Figure~\ref{fig:task32},~\ref{fig:CIFAR10_heatmap_appendix},~\ref{fig:CIFAR10_BC_diff_teacher},~\ref{fig:CIFAR10_B_student_aug},~\ref{fig:dataset_ipc},~\ref{fig:dataset_student_aug}, the total number of training steps is set to fixed 10,000 steps.

\paragraph{Model architecture.}
For the experiments on CIFAR10, CIFAR100, and TinyImageNet, both the teacher and student models were implemented using a modified ResNet-18 architecture (no pretrained weights) where the initial convolutional layer was replaced with a $3 \times 3$ convolutional layer, using stride 1 and padding 1 to better accommodate the input image sizes.
The batch normalization layer was adjusted accordingly to match the new convolutional output, and the max-pooling layer was removed.
To better evaluate the backbone of the model, we pruned the final fully connected layers, and a linear layer was attached from the output feature dimension to the number of classes in the corresponding dataset, the backbone of the model and the attached classifier layer were trained separately, as described earlier in~\secref{sec:definition}.

\paragraph{Pre-trained Teacher Models.}
To thoroughly investigate the influence of teacher models with varying accuracies on the representation capability of the student model $\mphi_{\mtheta}$, we first train a series of teacher models with differing accuracies, employing various augmentations on the training set.
using the traditional \SA strategy (i.e., using one-hot targets) and trained them with the standard cross-entropy loss function.
Validation was performed at the end of each step, and the teacher model was saved the first time it reached the preset accuracy, which was then used for subsequent student model training,
and we ensure that the validation error for each teacher model with a specific accuracy is less than 1\%.
For \SB and \SC, the student model is trained using the soft targets by the teacher model, with the temperature of the soft labels set to 2.0.

\paragraph{Plot settings.}
To reduce noise and provide clearer visualization of long-term trends, we applied locally weighted scatterplot smoothing (LOWESS)~\citep{cleveland1979robust} to the figures that examine long-term behavior, such as Figures~\ref{fig:task12},~\ref{fig:SBC},~\ref{fig:CIFAR10_BC_compare},~\ref{fig:dataset_exp1},~\ref{fig:dataset_exp2}, and~\ref{fig:dataset_teacher_diff_aug},
using 10\% of the data points for each local regression. Considering that the curve fitting applied is not well-suited for observing early-stage data, particularly when assessing the impact of \SC\ in the initial phases,
we provide additional numerical results in the early training phase in Table~\ref{tab:CIFAR10_compareBandC},~\ref{tab:dataset_compareBandC},~\ref{tab:ImageNet},~\ref{tab:ImageNet_AC}.
We conduct at least five trials for each experiment and report the standard deviations for the proposed three strategies,.

\paragraph{Hyper parameter settings.}
We employed the default implementation of the AdamW optimizer in the PyTorch Lightning Module, where the learning rate is set to $1 \times 10^{-3}$, and the weight decay is 0.01.
The metric for evaluating the performance of all models is their Top-1 validation accuracy on the test set. Experiments were conducted on NVIDIA RTX 4090 GPUs and Intel Xeon processors.

\paragraph{Augmentation definition.}
We define the data augmentation techniques mentioned throughout the paper as follows:

\textbf{No Augmentation (NoAug):} Only normalization is applied based on the corresponding dataset.

\textbf{Standard Augmentation (StdAug):}
\begin{itemize}
    \item \textbf{ColorJitter (or Color in short)}: Apply random changes to the brightness, contrast, saturation with a probability of 0.8 to each image.
    \item \textbf{RandomGrayscale}: Converts the image to grayscale with a 0.2 probability.
    \item \textbf{RandomHorizontalFlip}: Flips the image horizontally with a 50\% chance.
    \item \textbf{RandomResizedCrop (or Crop in short)}: Resizes and crops the image to the input size with a scaling factor between 0.08 and 1.0.
\end{itemize}

\textbf{Mix-based Augmentation (not include StdAug in our settings):}

\begin{itemize}
    \item \textbf{MixUp}: Perform a linear interpolation between two randomly selected images, where the interpolation factor, $\lambda$, is drawn from a Beta distribution with parameters $(\alpha, \beta) = (0.8, 0.8)$ which controls the degree of interpolation between the two images.
    \item \textbf{CutMix}: A bounding box is randomly generated within one image, and the pixel values inside the box are replaced by the corresponding region from a second randomly selected image. The area of the bounding box is determined by a parameter sampled from a Beta distribution with parameters $(\alpha, \beta) = (1.0, 1.0)$ which controls the relative size of the cutout region and the pasted-in region.
\end{itemize}

For all experiments except in ~\secref{sec:exp3},~\secref{sec:student_aug} and Tabel~\ref{tab:CIFAR10_diff_aug},~\ref{tab:dataset_teacher_aug}, Figure~\ref{fig:CIFAR10_heatmap_appendix},~\ref{fig:dataset_teacher_diff_aug},~\ref{fig:dataset_student_aug}, we employed the \textbf{Standard Augmentation (StdAug)} and normalization for both teacher models and student models.
For the experiments in Figure~\ref{fig:task31},~\ref{fig:dataset_teacher_diff_aug}, and~\autoref{tab:dataset_teacher_aug}, we applied \textbf{StdAug} to the student model.
For the experiments in Figure~\ref{fig:task5} and~\ref{fig:dataset_student_aug}, \textbf{StdAug} was applied to the teacher model.

\section{Additional Experimental Results} \label{app:results}
\begin{figure}[h!]
    \centering
    \includegraphics[width=.49\linewidth]{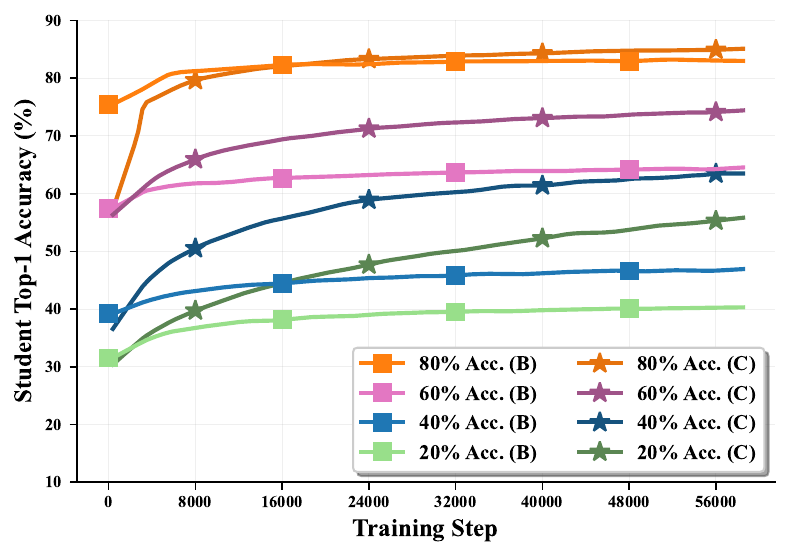}
    \caption{\textbf{Comparison of \SB\ and \SC\ under more different teacher models with different accuracies.} It further shows that as the accuracy of the teacher model increases, the advantage of using \SC\ over \SB\ becomes less apparent.}
    \label{fig:CIFAR10_BC_compare}
\end{figure}

We further conducted more experiments on the CIFAR10 dataset to supplement and verify the patterns observed in the main body.
We first show the experimental patterns can be observed more clearly under the step setting in~\secref{sec:exp1} through~\autoref{tab:CIFAR10_diff_step}.
And we compare \SB and \SC using more teachers during different training stages to observe under what conditions can \SC\ hold a greater advantage through~\autoref{tab:CIFAR10_compareBandC},
and we find that \SC will not yield a significant advantage when the accuracy of the teacher model is high enough (see~\autoref{fig:CIFAR10_BC_compare}).

In addition to the data augmentation strategies used in~\secref{sec:exp3}, we further compared the effects of applying more data augmentation strategies to teacher models with different accuracies in~\autoref{tab:CIFAR10_diff_aug}.
We find that in most cases only when applying the RandomResizedCrop strategy to the teacher model does it significantly help the training of the student model, while most other individual augmentation strategies tend to have a negative impact.

\begin{figure}[h!]
    \centering
    \begin{subfigure}{.49\textwidth}  %
        \centering
        \includegraphics[width=1.0\linewidth]{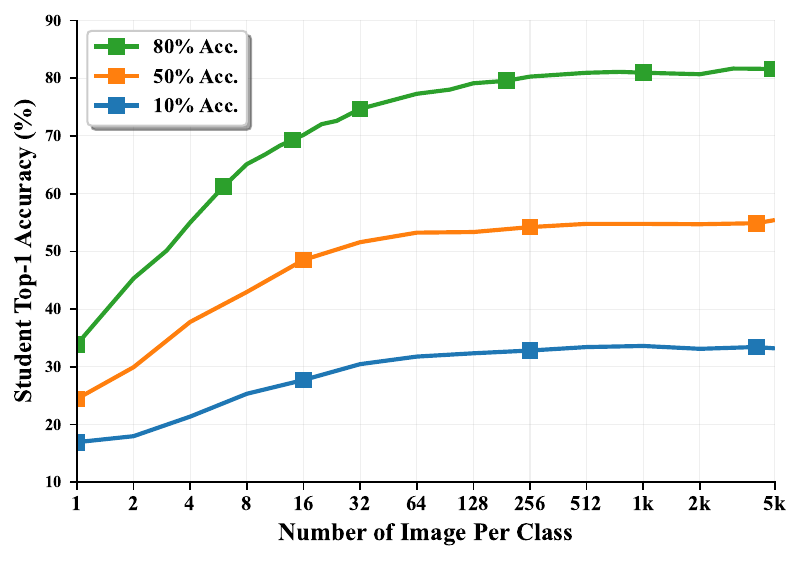}
        \captionsetup{labelformat=empty,skip=-8pt}
        \caption{}
        \label{fig:CIFAR10_SB_diff_teacher}
    \end{subfigure}
    \hfill %
    \begin{subfigure}{.49\textwidth}
        \centering
        \includegraphics[width=1.0\linewidth]{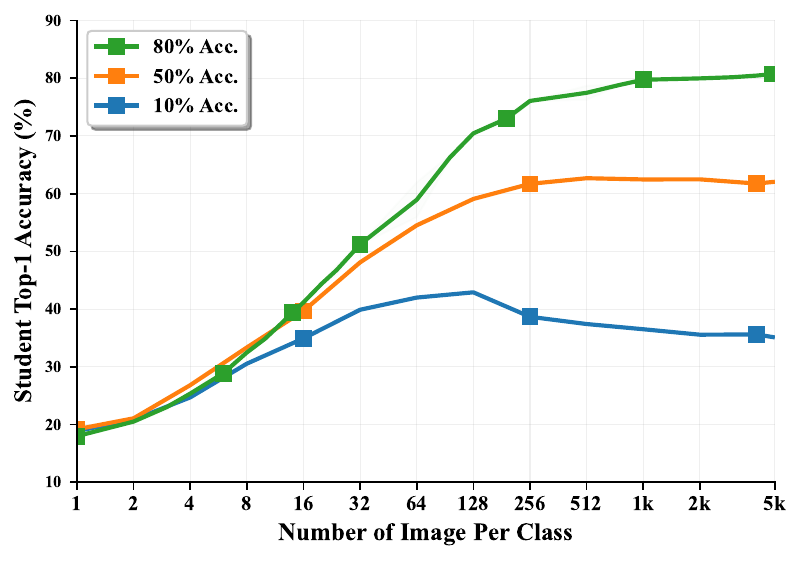}
        \captionsetup{labelformat=empty,skip=-8pt}
        \caption{}
        \label{fig:CIFAR10_SC_diff_teacher}
    \end{subfigure}
    \caption{\textbf{The scaling behavior across different teachers (by changing the number of image per class) under \SB (left) and \SC (right).} In \SB, different teacher models exhibit similar patterns of scaling, whereas in \SC, the performance of the student model trained by a randomly initialized teacher model (i.e., 10\% Acc.) initially improves but then declines as the IPC increases.}
    \label{fig:CIFAR10_BC_diff_teacher}
\end{figure}

\begin{figure}[h!]
    \centering
    \begin{subfigure}{.32\textwidth}  %
        \centering
        \includegraphics[width=1.0\linewidth]{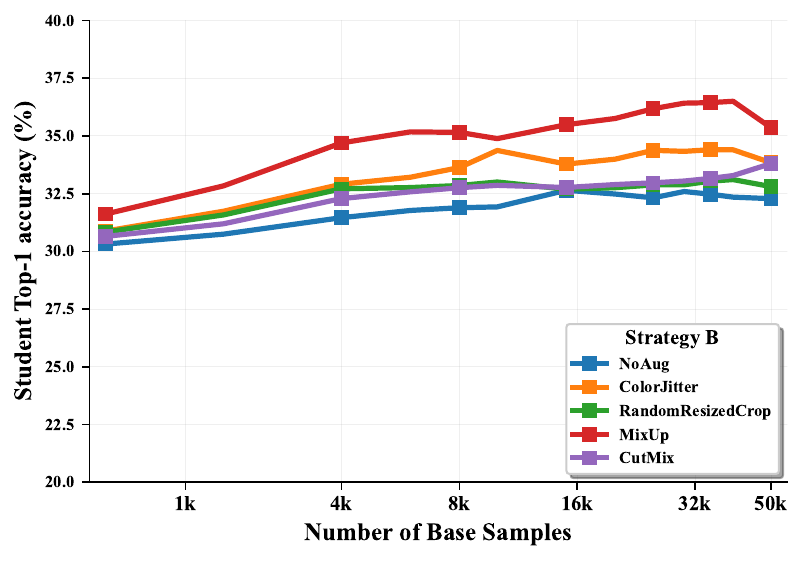}
        \captionsetup{labelformat=empty,skip=-8pt}
        \caption{}
        \label{fig:CIFAR10_B_10_aug}
    \end{subfigure}
    \hfill %
    \begin{subfigure}{.32\textwidth}
        \centering
        \includegraphics[width=1.0\linewidth]{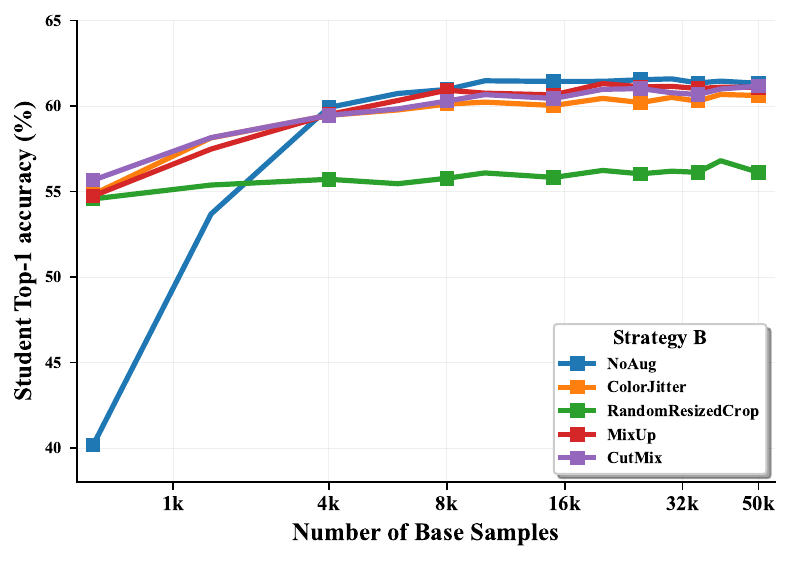}
        \captionsetup{labelformat=empty,skip=-8pt}
        \caption{}
        \label{fig:CIFAR10_B_50_aug}
    \end{subfigure}
    \hfill %
    \begin{subfigure}{.32\textwidth}
        \centering
        \includegraphics[width=1.0\linewidth]{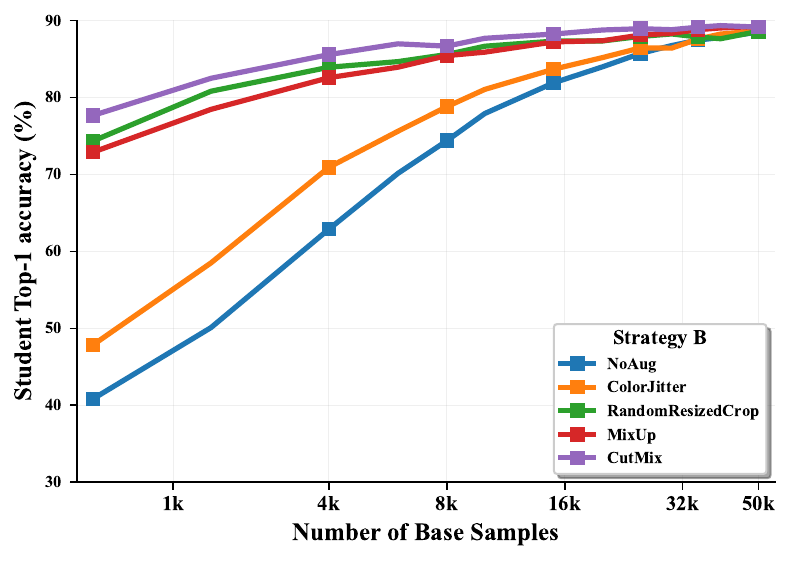}
        \captionsetup{labelformat=empty,skip=-8pt}
        \caption{}
        \label{fig:CIFAR10_B_90_aug}
    \end{subfigure}
    \caption{\textbf{The scaling behavior across different teacher using different data augmentation under \SB.} From left to right, the teacher models have accuracies of 10\%, 50\%, and 90\%, respectively. For a teacher model with moderate or high accuracy, applying different data augmentation strategies to the student model results in their performance converging to nearly the same value (except under the 50\% teacher model where using the RandomResizedCrop strategy for the student model is insensitive to changes in the number of samples). However, for a low-accuracy teacher model (in this case, a completely random initialized teacher model), applying different data augmentation strategies to the student model shows slight variations in effect as the number of samples increases.}
    \label{fig:CIFAR10_B_student_aug}
\end{figure}

Aside from the investigation of samples in~\secref{sec:sample}, we further conducted more experiments and found that the scaling behavior of student models trained by teacher models with different accuracies shows slight differences, as shown in Figure~\ref{fig:CIFAR10_BC_diff_teacher} and~\ref{fig:CIFAR10_B_student_aug}.

\begin{figure}[h!]
    \centering
    \begin{subfigure}{0.3\textwidth}
        \centering
        \includegraphics[width=1.0\linewidth]{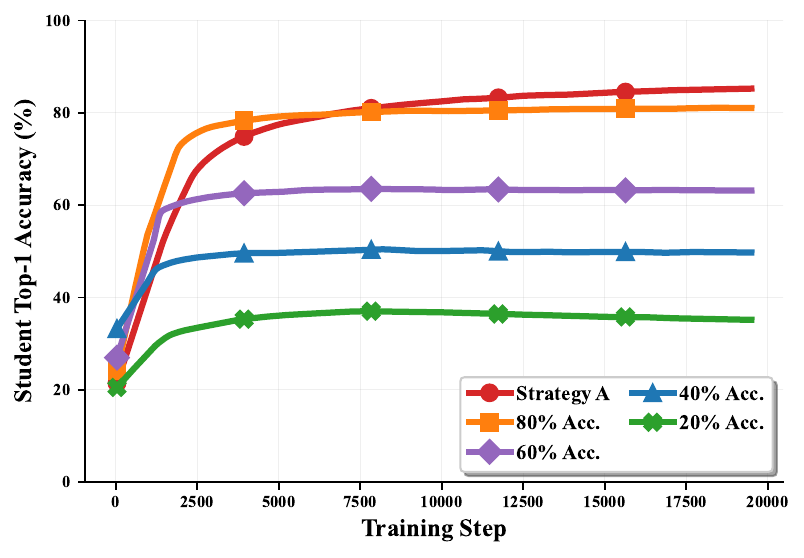}
        \caption{\centering batch size = 128, learning rate = 1e-2}
    \end{subfigure}
    \hfill
    \begin{subfigure}{0.3\textwidth}
        \centering
        \includegraphics[width=1.0\linewidth]{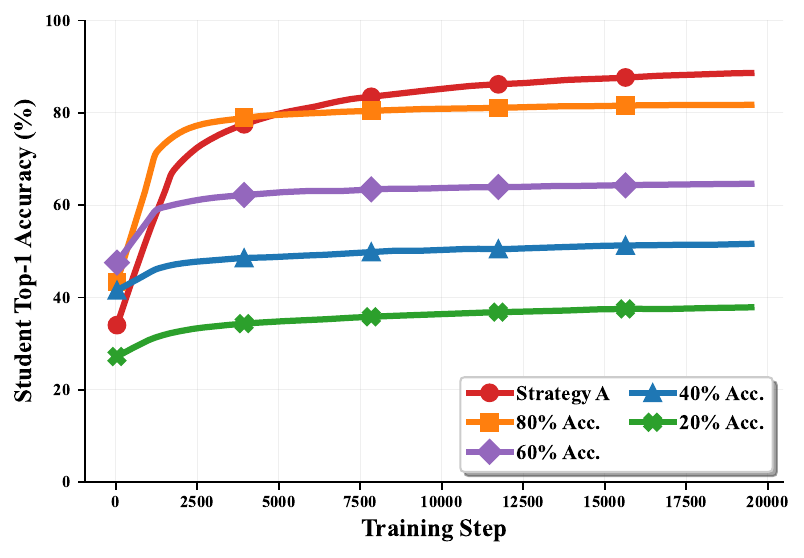}
        \caption{\centering batch size = 128, learning rate = 1e-3}
    \end{subfigure}
    \hfill
    \begin{subfigure}{0.3\textwidth}
        \centering
        \includegraphics[width=1.0\linewidth]{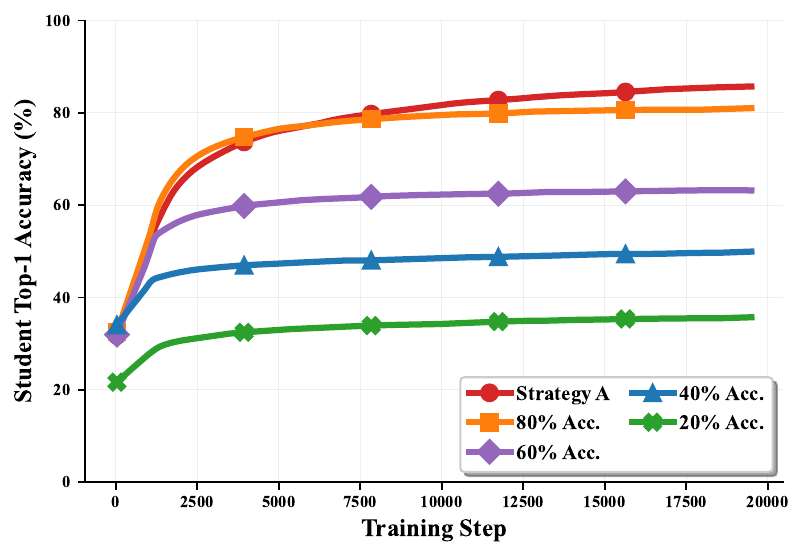}
        \caption{\centering batch size = 128, learning rate = 1e-4}
    \end{subfigure}

    \vspace{0.5cm} %
    \begin{subfigure}{0.3\textwidth}
        \centering
        \includegraphics[width=1.0\linewidth]{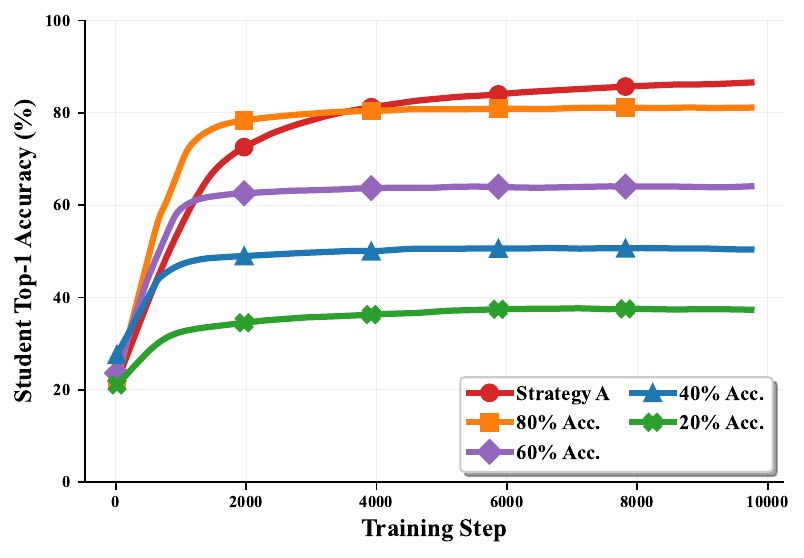}
        \caption{\centering batch size = 256, learning rate = 1e-2}
    \end{subfigure}
    \hfill
    \begin{subfigure}{0.3\textwidth}
        \centering
        \includegraphics[width=1.0\linewidth]{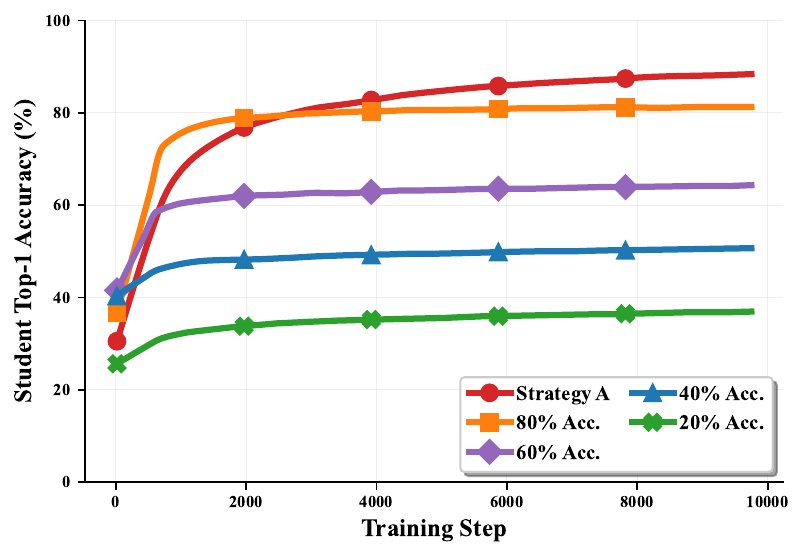}
        \caption{\centering batch size = 256, learning rate = 1e-3}
    \end{subfigure}
    \hfill
    \begin{subfigure}{0.3\textwidth}
        \centering
        \includegraphics[width=1.0\linewidth]{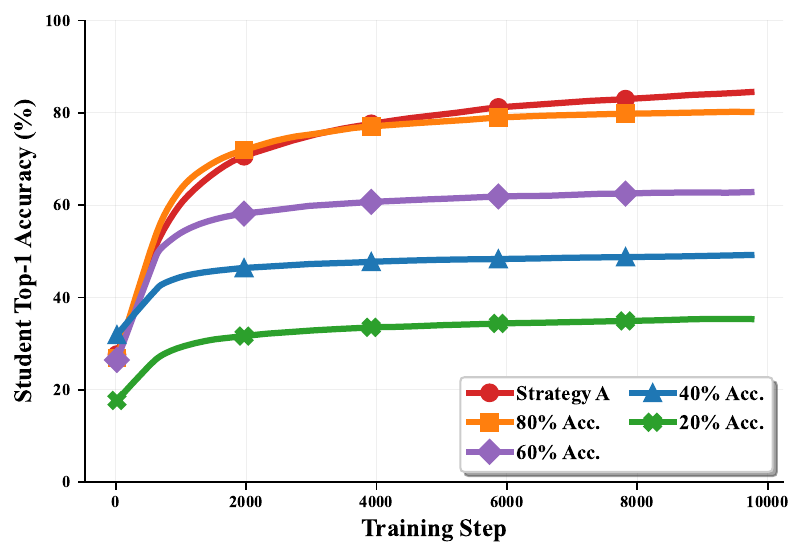}
        \caption{\centering batch size = 256, learning rate = 1e-4}
    \end{subfigure}
    \caption{\textbf{The comparison of strategies \SA and \SB under different learning rate and batch size settings.} All results show that the \SB has a short term advantage while \SA have a long term advantage. }
    \label{fig:ablation}
\end{figure}

In addition to the experiments above, we performed ablation studies on the effects of batch size and learning rate, as shown in Figure~\ref{fig:ablation} and Table~\ref{tab:ablation}.

All values presented in the tables represent the Top-1 accuracy of the student model under the corresponding teacher model settings with the percentage symbol (\%) omitted for simplicity.

\begin{table}[ht]
    \centering
    \caption{\textbf{The Top-1 accuracy of the student model under different training steps and different teachers.} The \textbf{bolded} data represent the maximum student Top-1 accuracy in each training step. Only under a limited number of training steps (i.e., 1000 steps setting in the main body), the teacher model with relatively medium accuracy will show its advantage.}
    \renewcommand{\arraystretch}{1.2}
    \begin{tabular}{cccccccccc}
        \toprule
        \multirow{2}{*}{Training Step} & \multicolumn{9}{c}{Teacher Top-1 Accuracy (\%) $\pm$1\%}                                                                                                     \\
        \cmidrule(lr){2-10}
                                       & 10\%                                                     & 20\%  & 30\%  & 40\%  & 50\%           & 60\%           & 70\%           & 80\%           & 90\%  \\
        \midrule
        10                             & 10.59                                                    & 12.92 & 13.39 & 14.08 & 13.86          & \textbf{15.53} & 14.25          & 15.14          & 14.38 \\
        20                             & 11.85                                                    & 18.57 & 18.79 & 18.47 & \textbf{19.51} & 19.02          & 19.11          & 17.94          & 19.37 \\
        50                             & 13.46                                                    & 20.94 & 23.54 & 24.97 & \textbf{27.69} & 26.64          & 27.46          & 26.42          & 25.52 \\
        100                            & 14.76                                                    & 23.19 & 26.71 & 31.15 & \textbf{35.24} & 32.55          & 32.32          & 33.00          & 32.22 \\
        150                            & 15.63                                                    & 25.59 & 27.95 & 33.38 & 37.77          & \textbf{37.81} & 36.57          & 36.52          & 34.66 \\
        200                            & 18.52                                                    & 26.23 & 31.11 & 36.10 & 43.29          & \textbf{43.31} & 41.32          & 39.44          & 37.93 \\
        300                            & 19.16                                                    & 27.66 & 32.00 & 36.47 & 46.73          & \textbf{51.51} & 50.69          & 46.67          & 42.99 \\
        500                            & 19.49                                                    & 29.26 & 33.43 & 38.06 & 48.93          & 55.72          & \textbf{61.06} & 57.68          & 54.79 \\
        700                            & 21.99                                                    & 29.84 & 33.59 & 38.73 & 48.83          & 56.50          & 63.71          & \textbf{65.16} & 62.27 \\
        1000                           & 24.38                                                    & 31.17 & 33.57 & 39.32 & 49.63          & 58.38          & 65.91          & \textbf{70.13} & 67.90 \\
        1200                           & 24.50                                                    & 32.35 & 34.98 & 40.64 & 51.56          & 58.81          & 67.45          & \textbf{73.09} & 70.23 \\
        1500                           & 25.88                                                    & 32.30 & 35.31 & 39.92 & 51.76          & 58.79          & 68.00          & \textbf{75.29} & 73.92 \\
        \bottomrule
    \end{tabular}
    \label{tab:CIFAR10_diff_step}
\end{table}

\begin{table}[ht]
    \centering
    \caption{\textbf{Comparing the student models trained with \SB and \SC at different training stages.} The Gain represents the performance improvement factor of the student model using \SC compared to \SB, with \textbf{bold} represent cases where the gain is greater than 1 (i.e., \SC is more advantageous). For a high-accuracy teacher model, \SC does not show advantages with fewer training steps. However, for a low-accuracy teacher model, training advantages are observed in both early and late training stages. And as the accuracy of the teacher model increases from 10\% to 90\%, the gain of using \SC\ compared to \SB\ first increases then gradually decreases.}
    \renewcommand{\arraystretch}{1.2}
    \setlength{\tabcolsep}{3pt}
    \resizebox{\textwidth}{!}{
    \begin{tabular}{cc|ccccccccc}
        \toprule
        \multirow{2}{*}{Step} & \multirow{2}{*}{Strategy} & \multicolumn{9}{c}{Teacher Top-1 Accuracy (\%) $\pm$1\%}                                                                                                                         \\
        \cmidrule(lr){3-11}
                              &                           & 10\%                        & 20\%                        & 30\%                        & 40\%                        & 50\%                        & 60\%                        & 70\%                        & 80\%                        & 90\%  \\
        \midrule
        \multirow{3}{*}{1950}
        & B                         & 25.09 {\scriptsize$\pm$ 0.5} & 33.31 {\scriptsize$\pm$ 0.6} & 35.93 {\scriptsize$\pm$ 0.4} & 40.29 {\scriptsize$\pm$ 0.6} & 51.63 {\scriptsize$\pm$ 0.3} & 59.12 {\scriptsize$\pm$ 0.6} & 66.93 {\scriptsize$\pm$ 0.4} & 73.31 {\scriptsize$\pm$ 0.4} & 72.80 {\scriptsize$\pm$ 0.6} \\
        & C                         & 27.06 {\scriptsize$\pm$ 0.7} & 32.54 {\scriptsize$\pm$ 0.9} & 34.79 {\scriptsize$\pm$ 0.7} & 41.28 {\scriptsize$\pm$ 0.5} & 52.95 {\scriptsize$\pm$ 0.4} & 57.55 {\scriptsize$\pm$ 0.7} & 63.14 {\scriptsize$\pm$ 0.6} & 67.99 {\scriptsize$\pm$ 0.5} & 68.66 {\scriptsize$\pm$ 0.5} \\
        & Gain                      & \textbf{1.08}               & 0.98                        & 0.97                        & \textbf{1.02}               & \textbf{1.03}               & 0.97                        & 0.94                        & 0.93                        & 0.94  \\
  \midrule
  \multirow{3}{*}{7800}
        & B                         & 32.02 {\scriptsize$\pm$ 0.4} & 36.66 {\scriptsize$\pm$ 0.5} & 38.35 {\scriptsize$\pm$ 0.5} & 43.19 {\scriptsize$\pm$ 0.4} & 54.44 {\scriptsize$\pm$ 0.4} & 61.76 {\scriptsize$\pm$ 0.7} & 71.24 {\scriptsize$\pm$ 0.6} & 81.27 {\scriptsize$\pm$ 0.6} & 87.33 {\scriptsize$\pm$ 0.4} \\
        & C                         & 34.07 {\scriptsize$\pm$ 0.5} & 39.86 {\scriptsize$\pm$ 0.8} & 42.96 {\scriptsize$\pm$ 0.5} & 50.57 {\scriptsize$\pm$ 0.6} & 60.68 {\scriptsize$\pm$ 0.5} & 65.89 {\scriptsize$\pm$ 0.6} & 72.47 {\scriptsize$\pm$ 0.5} & 79.67 {\scriptsize$\pm$ 0.7} & 84.00 {\scriptsize$\pm$ 0.6} \\
        & Gain                      & \textbf{1.06}               & \textbf{1.09}               & \textbf{1.12}               & \textbf{1.17}               & \textbf{1.12}               & \textbf{1.07}               & \textbf{1.02}               & 0.98                        & 0.96  \\
  \midrule
  \multirow{3}{*}{39000}
        & B                         & 35.96 {\scriptsize$\pm$ 0.4} & 39.70 {\scriptsize$\pm$ 0.4} & 41.61 {\scriptsize$\pm$ 0.5} & 46.11 {\scriptsize$\pm$ 0.5} & 57.18 {\scriptsize$\pm$ 0.3} & 63.90 {\scriptsize$\pm$ 0.4} & 72.58 {\scriptsize$\pm$ 0.5} & 82.93 {\scriptsize$\pm$ 0.6} & 90.35 {\scriptsize$\pm$ 0.4} \\
        & C                         & 42.86 {\scriptsize$\pm$ 0.5} & 51.98 {\scriptsize$\pm$ 0.2} & 55.96 {\scriptsize$\pm$ 0.4} & 61.34 {\scriptsize$\pm$ 0.6} & 69.84 {\scriptsize$\pm$ 0.4} & 73.02 {\scriptsize$\pm$ 0.5} & 77.93 {\scriptsize$\pm$ 0.6} & 84.29 {\scriptsize$\pm$ 0.5} & 91.17 {\scriptsize$\pm$ 0.5} \\
        & Gain                      & \textbf{1.19}                                            & \textbf{1.31} & \textbf{1.35} & \textbf{1.33} & \textbf{1.22} & \textbf{1.14} & \textbf{1.07} & \textbf{1.02} & \textbf{1.01}  \\
        \bottomrule
    \end{tabular}
    }
    \label{tab:CIFAR10_compareBandC}
\end{table}

\begin{table}[ht]
    \centering
    \caption{\textbf{Comparison of different augmentation method applied for the teacher model using \SB (training step=7800).} The \textbf{bolded} data represent the maximum student Top-1 accuracy in each column. Besides the advantage of NoAug over MixUp in the teacher model shown in section~\ref{sec:exp3}, this table further indicates that applying RandomResizedCrop to the teacher model will benefit the training of the student model, regardless of the accuracy of the teacher model.}
    \renewcommand{\arraystretch}{1.2}
    \begin{tabular}{ccccccccc}
        \toprule
        \multirow{2}{*}{\shortstack{Augmentation                                                                                                            \\ (on teacher)}} & \multicolumn{8}{c}{Teacher Top-1 Accuracy (\%) $\pm$1\%} \\
        \cmidrule(lr){2-9}
                    & 10\%           & 20\%           & 30\%           & 40\%           & 50\%           & 60\%           & 70\%           & 80\%           \\
        \midrule
        NoAug       & 31.20          & 35.55          & 40.29          & 43.92          & 49.50          & 56.94          & 67.59          & 77.43          \\
        ColorJitter & \textbf{32.18} & 35.74          & 38.44          & 41.61          & 50.06          & 58.83          & 65.88          & 77.37          \\
        Crop        & 31.14          & \textbf{37.43} & \textbf{41.27} & 42.93          & \textbf{54.26} & \textbf{66.11} & \textbf{75.37} & \textbf{82.28} \\
        MixUp       & 30.30          & 35.00          & 39.81          & 43.79          & 50.93          & 61.44          & 65.79          & 75.78          \\
        CutMix      & 30.74          & 35.86          & 37.78          & 41.23          & 48.41          & 55.72          & 68.65          & 73.24          \\
        StdAug      & 31.53          & 35.65          & 38.49          & \textbf{49.52} & 53.57          & 63.34          & 73.00          & 80.59          \\
        \bottomrule
    \end{tabular}
    \label{tab:CIFAR10_diff_aug}
\end{table}

\begin{table}[ht]
    \centering
    \caption{\textbf{The results of student models trained with different batch sizes and learning rates under \SA and \SB, evaluated across varying levels of teacher model accuracy.} For each configuration, we report the mean Top-1 accuracy of the student model. The results are consistent for each setting.}
    \renewcommand{\arraystretch}{1.2}
    \begin{tabular}{cc|ccccc}
        \toprule
        \multirow{2}{*}{batch size} & \multirow{2}{*}{\shortstack{learning                                                                                           \\ rate}} & \multicolumn{5}{c}{Teacher Top-1 Accuracy (\%) $\pm$1\%} \\
        \cmidrule(lr){3-7}
                                    &                                      & 20\%            & 40\%            & 60\%            & 80\%            & NoTeach         \\
        \midrule
        \multirow{3}{*}{128}        & 1e-2                                 & 35.42 $\pm$ 0.5 & 49.54 $\pm$ 0.9 & 63.23 $\pm$ 0.5 & 81.18 $\pm$ 0.4 & 85.52 $\pm$ 0.5 \\
        & 1e-3                                 & 37.86 $\pm$ 0.3 & 51.44 $\pm$ 0.3 & 64.54 $\pm$ 0.8 & 81.68 $\pm$ 0.3 & 88.59 $\pm$ 0.5 \\
        & 1e-4                                 & 34.88 $\pm$ 0.5 & 49.63 $\pm$ 0.5 & 63.20 $\pm$ 0.7 & 81.09 $\pm$ 0.3 & 86.01 $\pm$ 0.4 \\
\midrule
\multirow{3}{*}{256}        & 1e-2                                 & 37.39 $\pm$ 0.3 & 50.70 $\pm$ 0.4 & 64.42 $\pm$ 0.6 & 81.78 $\pm$ 0.4 & 86.50 $\pm$ 0.7 \\
        & 1e-3                                 & 37.06 $\pm$ 0.6 & 50.35 $\pm$ 0.4 & 64.29 $\pm$ 0.5 & 81.69 $\pm$ 0.4 & 88.53 $\pm$ 0.2 \\
        & 1e-4                                 & 34.88 $\pm$ 0.5 & 48.66 $\pm$ 0.4 & 63.15 $\pm$ 0.5 & 76.73 $\pm$ 8.2 & 84.42 $\pm$ 0.6 \\
        \bottomrule
    \end{tabular}
    \label{tab:ablation}
\end{table}

\section{Additional Dataset} \label{app:diff_datasets}
\begin{figure}[h!]
    \centering
    \begin{subfigure}{.49\textwidth}  %
        \centering
        \includegraphics[width=1.0\linewidth]{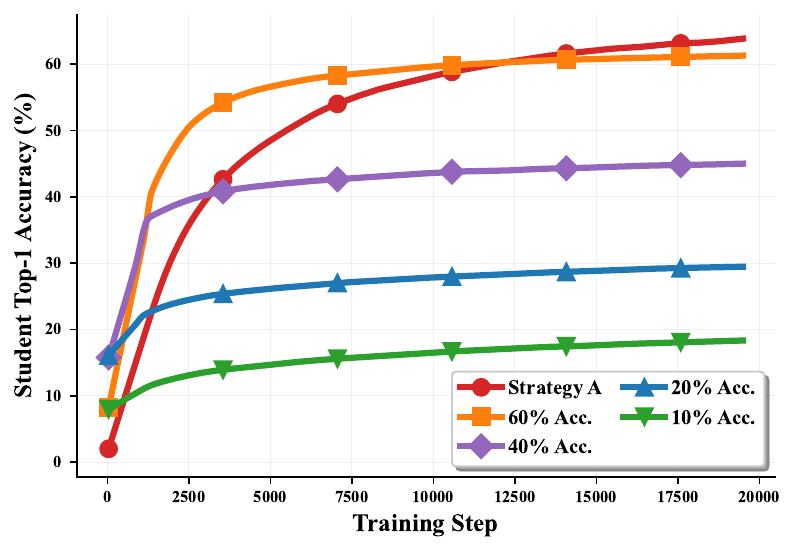}
        \captionsetup{labelformat=empty,skip=-8pt}
        \caption{}
        \label{fig:CIFAR100_exp1}
    \end{subfigure}
    \hfill %
    \begin{subfigure}{.49\textwidth}
        \centering
        \includegraphics[width=1.0\linewidth]{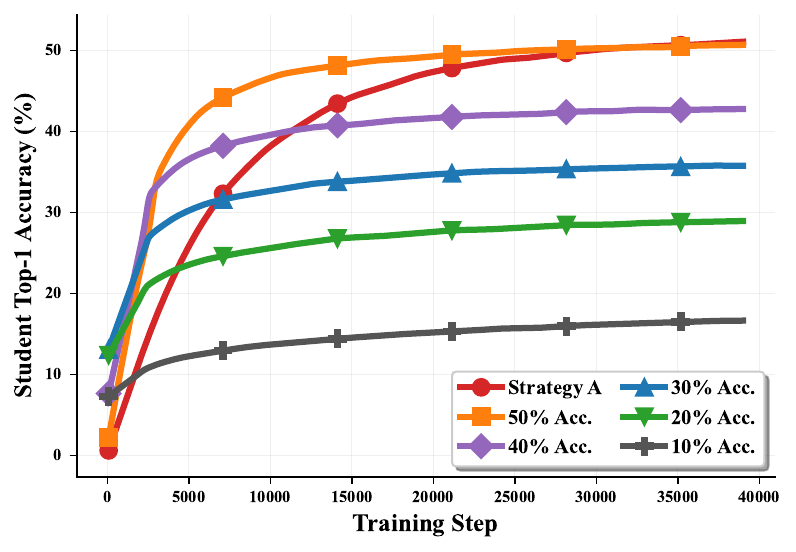}
        \captionsetup{labelformat=empty,skip=-8pt}
        \caption{}
        \label{fig:TinyImageNet_exp1}
    \end{subfigure}
    \caption{\textbf{Comparison of \SA and \SB using different teachers on CIFAR100 (Left) and TinyImageNet (Right).} Both show the long-term advantages of \SA, and the weaker teacher models exhibit more significant improvements in the performance of the student models.}
    \label{fig:dataset_exp1}
\end{figure}

\begin{figure}[h!]
    \centering
    \begin{subfigure}{.49\textwidth}  %
        \centering
        \includegraphics[width=1.0\linewidth]{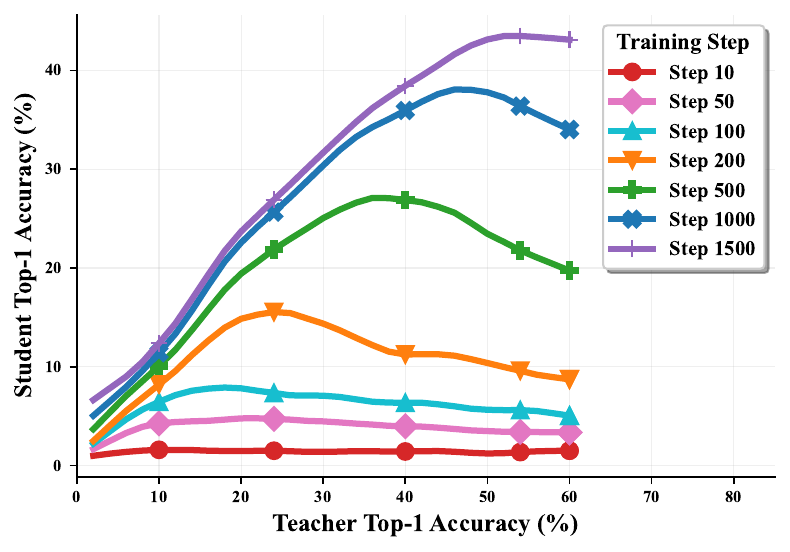}
        \captionsetup{labelformat=empty,skip=-8pt}
        \caption{}
        \label{fig:CIFAR100_exp2}
    \end{subfigure}
    \hfill %
    \begin{subfigure}{.49\textwidth}
        \centering
        \includegraphics[width=1.0\linewidth]{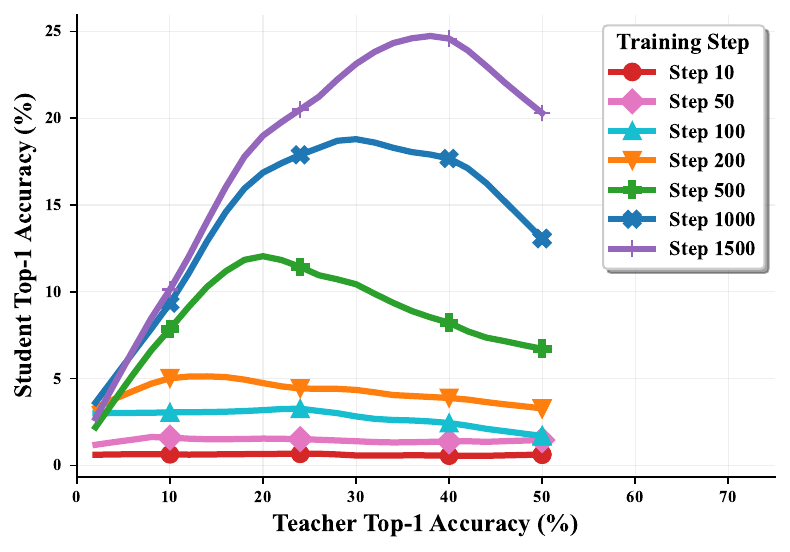}
        \captionsetup{labelformat=empty,skip=-8pt}
        \caption{}
        \label{fig:TinyImageNet_exp2}
    \end{subfigure}
    \caption{\textbf{The different behavior across teachers in different early-stage training steps on CIFAR100 (Left) and TinyImageNet (Right).} For a teacher model with moderate accuracy, there is an advantage in the early stages of training the student model.}
    \label{fig:dataset_exp2}
\end{figure}

\begin{figure}[h!]
    \centering
    \begin{subfigure}{.49\textwidth}  %
        \centering
        \includegraphics[width=1.0\linewidth]{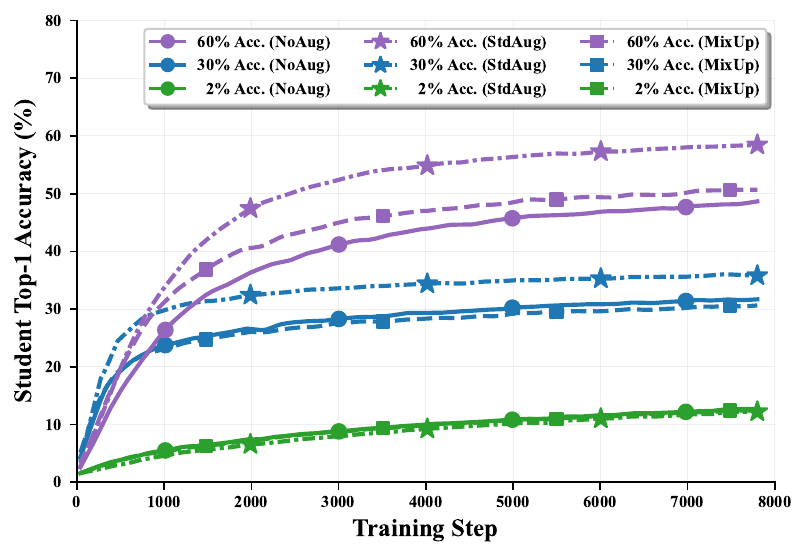}
        \captionsetup{labelformat=empty,skip=-8pt}
        \caption{}
        \label{fig:CIFAR100_teacher_diff_aug}
    \end{subfigure}
    \hfill %
    \begin{subfigure}{.49\textwidth}
        \centering
        \includegraphics[width=1.0\linewidth]{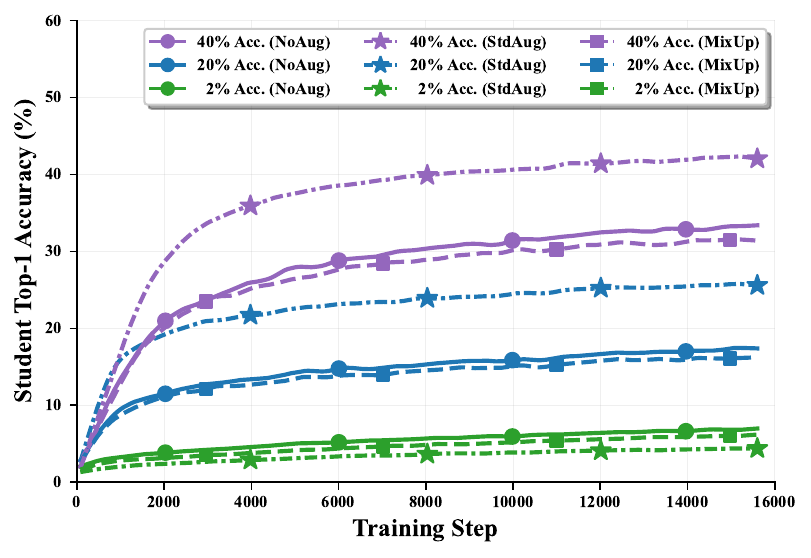}
        \captionsetup{labelformat=empty,skip=-8pt}
        \caption{}
        \label{fig:TinyImageNet_teacher_diff_aug}
    \end{subfigure}
    \caption{\textbf{The different impacts on student model training when applying different image augmentation strategies to the teacher model under \SB, with the left figure representing CIFAR100 and the right figure representing TinyImageNet.} As the accuracy of the teacher model increases, aggressive data augmentation strategies like MixUp show a stronger advantage. However, when the accuracy of the teacher model is low, MixUp actually exhibits a more pronounced disadvantage.}
    \label{fig:dataset_teacher_diff_aug}
\end{figure}

\begin{figure}[h!]
    \centering
    \begin{subfigure}{.49\textwidth}  %
        \centering
        \includegraphics[width=1.0\linewidth]{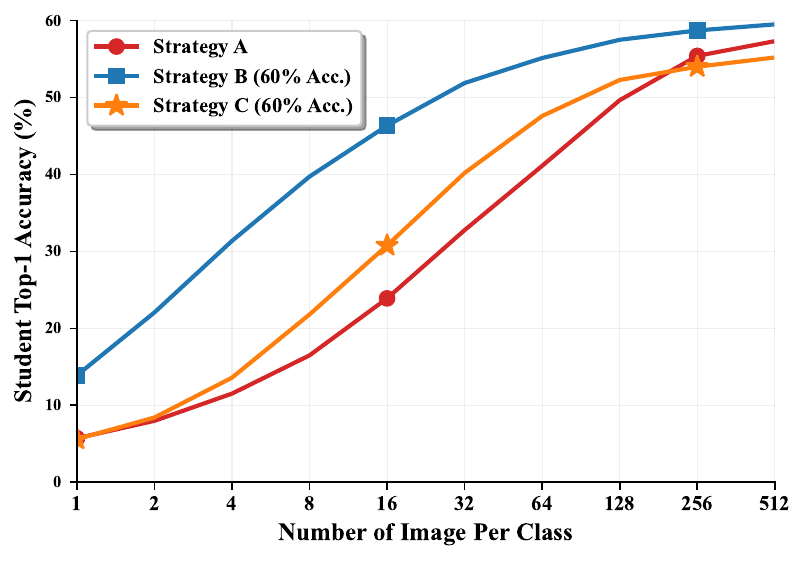}
        \captionsetup{labelformat=empty,skip=-8pt}
        \caption{}
        \label{fig:CIFAR100_ipc}
    \end{subfigure}
    \hfill %
    \begin{subfigure}{.49\textwidth}
        \centering
        \includegraphics[width=1.0\linewidth]{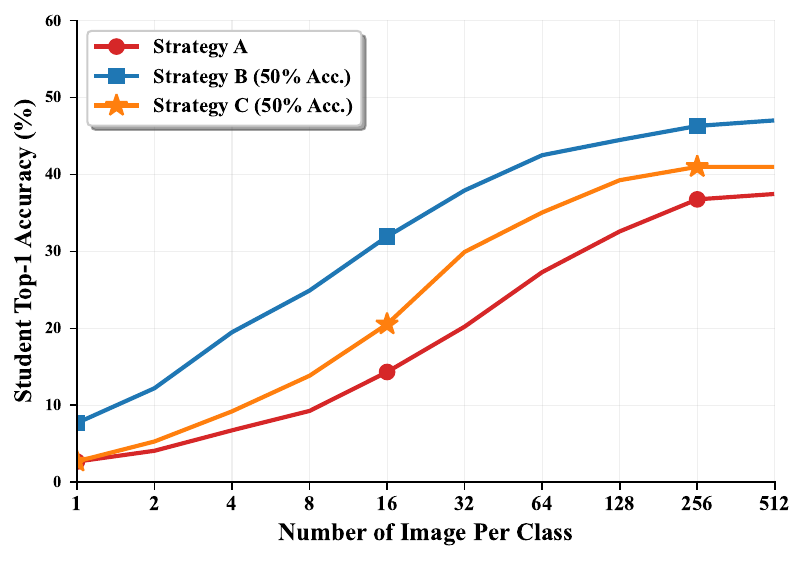}
        \captionsetup{labelformat=empty,skip=-8pt}
        \caption{}
        \label{fig:TinyImageNet_ipc}
    \end{subfigure}
    \caption{\textbf{The scaling behavior using different strategy for training on CIFAR100 (Left) and TinyImageNet (Right).} In both datasets, \SB\ consistently shows a greater advantage, \SC has some advantage when the sample size is small. And for CIFAR100, when the sample size is sufficient enough  (i.e., using all samples where IPC=500), the performance of \SC is eventually surpassed by \SA.}
    \label{fig:dataset_ipc}
\end{figure}

\begin{figure}[h!]
    \centering
    \begin{subfigure}{0.3\textwidth}
        \centering
        \includegraphics[width=1.0\linewidth]{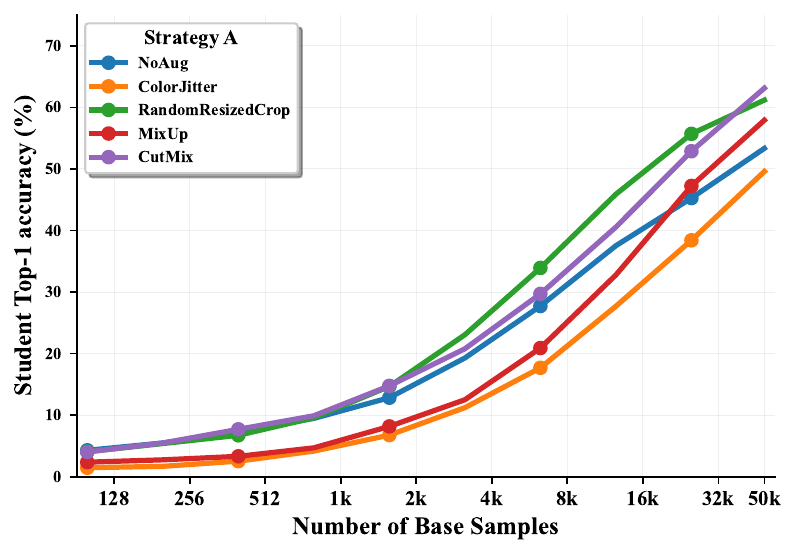}
        \captionsetup{labelformat=empty,skip=-8pt}
        \caption{}
    \end{subfigure}
    \hfill
    \begin{subfigure}{0.3\textwidth}
        \centering
        \includegraphics[width=1.0\linewidth]{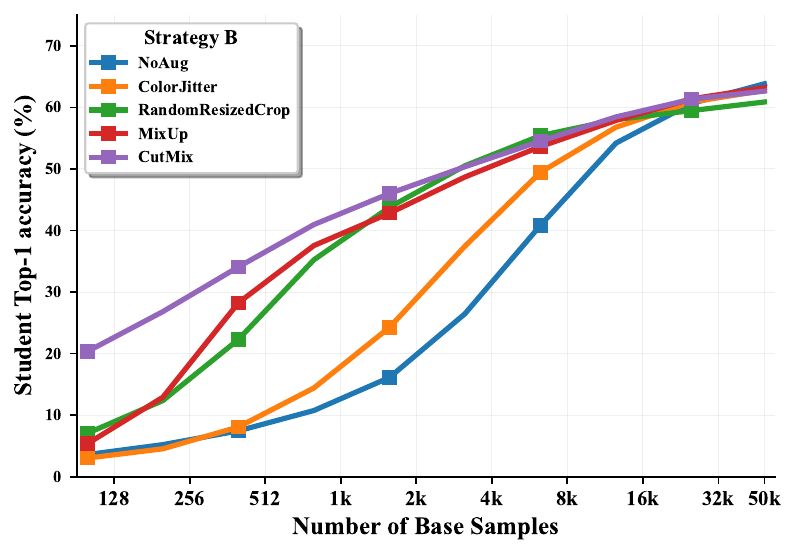}
        \captionsetup{labelformat=empty,skip=-8pt}
        \caption{}
    \end{subfigure}
    \hfill
    \begin{subfigure}{0.3\textwidth}
        \centering
        \includegraphics[width=1.0\linewidth]{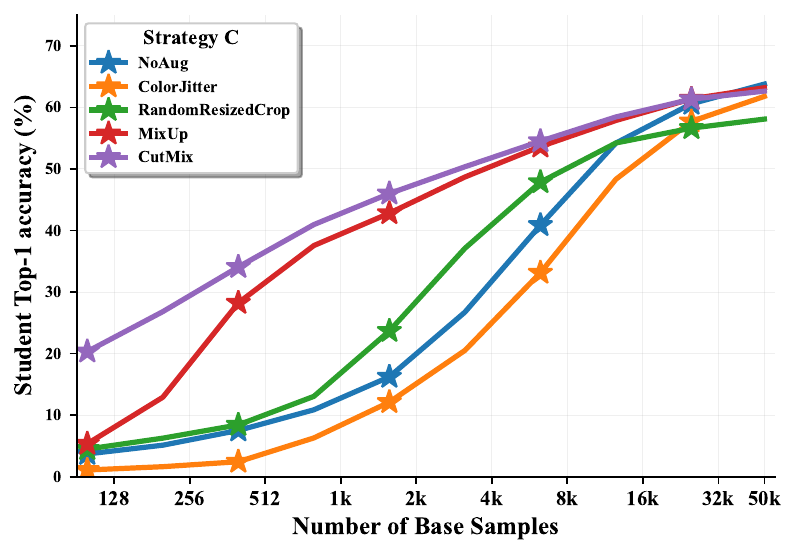}
        \captionsetup{labelformat=empty,skip=-8pt}
        \caption{}
    \end{subfigure}

    \vspace{0.5cm} %
    \begin{subfigure}{0.3\textwidth}
        \centering
        \includegraphics[width=1.0\linewidth]{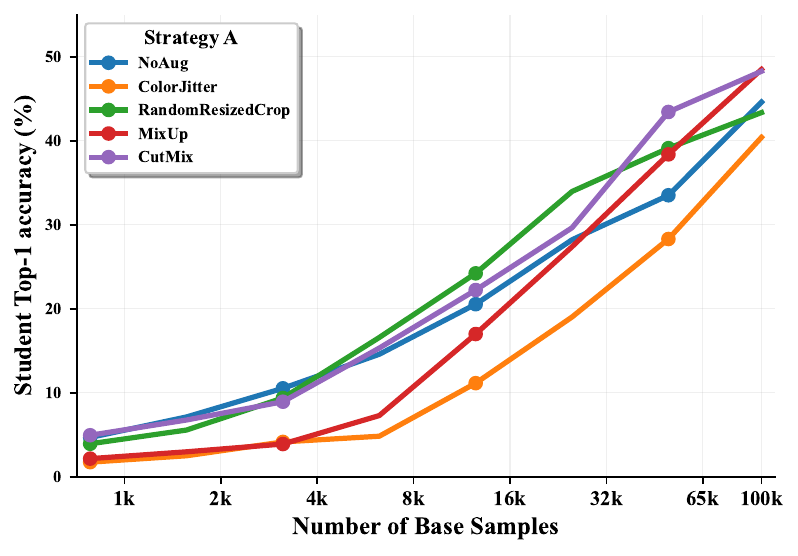}
        \captionsetup{labelformat=empty,skip=-8pt}
        \caption{}
    \end{subfigure}
    \hfill
    \begin{subfigure}{0.3\textwidth}
        \centering
        \includegraphics[width=1.0\linewidth]{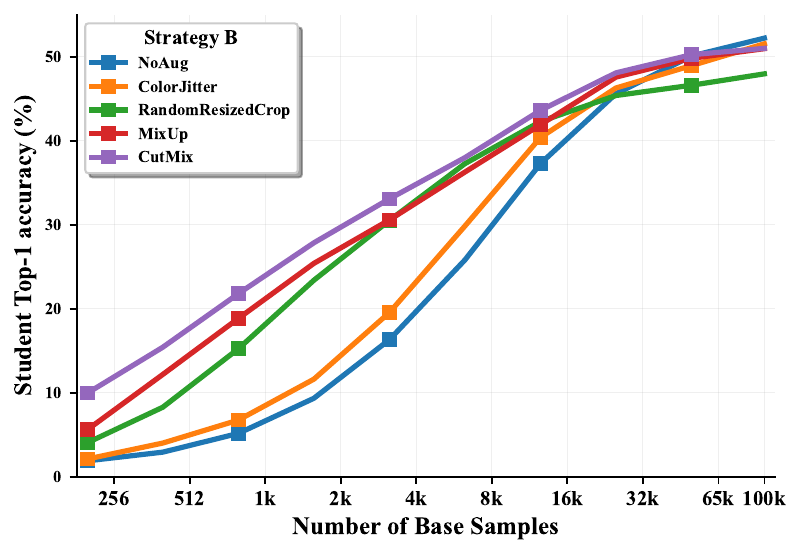}
        \captionsetup{labelformat=empty,skip=-8pt}
        \caption{}
    \end{subfigure}
    \hfill
    \begin{subfigure}{0.3\textwidth}
        \centering
        \includegraphics[width=1.0\linewidth]{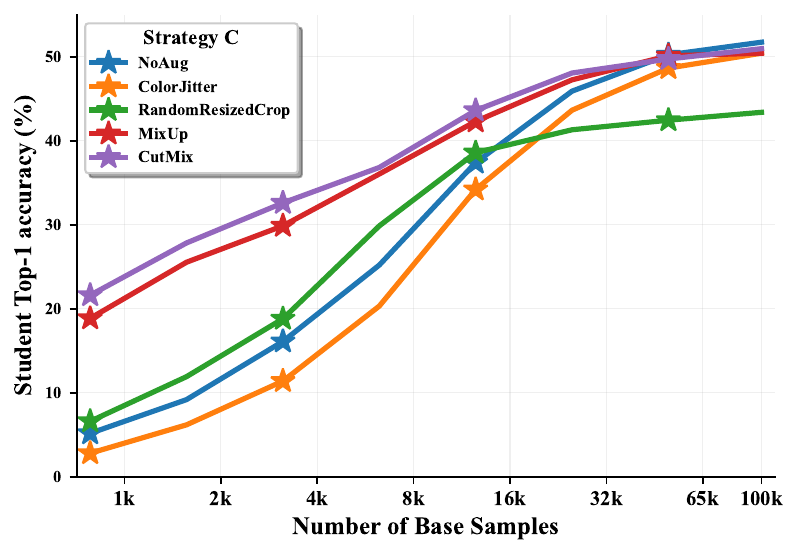}
        \captionsetup{labelformat=empty,skip=-8pt}
        \caption{}
    \end{subfigure}
    \caption{\textbf{The scaling behavior of the student model using different image augmentation methods under the three strategies.} The three images above show the results of CIFAR100, while the three images below show the results of TinyImageNet, all of which demonstrate the advantages of \SB and \SC compared to the traditional \SA under limited samples.}
    \label{fig:dataset_student_aug}
\end{figure}

We further conducted experiments on different datasets, including CIFAR100 and TinyImageNet to validate the findings presented in the main body.
We first compared \SA\ and \SB\ on the two datasets in Figure~\ref{fig:dataset_exp1} and \ref{fig:dataset_exp2}, and further investigated the behavior of \SC\ through \autoref{tab:dataset_compareBandC}.
We also compared the impact of applying different image augmentation strategies to the teacher models in Figure~\ref{fig:dataset_teacher_diff_aug} and \autoref{tab:dataset_teacher_aug},
and further explored the scaling behavior influenced by changing the sample size and applying different augmentation strategies on the student models, as shown in Figure~\ref{fig:dataset_ipc} and \ref{fig:dataset_student_aug}.
To have a more comprehensive evaluation on the differences between our proposed strategy \SC\ and traditional methods,
we conducted experiments on the ImageNet dataset, which is larger and more representative of real-world scenarios.
We selected ResNet50 and ViT as the backbone architectures for teacher and student models, and batch size was set to 512, and we validated every 500 steps, with other hyperparameters remian unchanged. The visual results are shown in Figure~\ref{fig:ImageNet}, while detailed numerical results are provided in Table~\ref{tab:ImageNet},~\ref{tab:ImageNet_AC}.

\begin{table}[ht]
    \centering
    \caption{\textbf{Comparing the student models trained with \SB and \SC on different datasets. } Both show that under \SC, teacher models with moderate or low accuracy are more beneficial for improving the performance of student models.}
    \renewcommand{\arraystretch}{1.2}
    \setlength{\tabcolsep}{3pt}
    \resizebox{\textwidth}{!}{
    \begin{tabular}{ccc|ccccccc}
        \toprule
        \multirow{2}{*}{Dataset} & \multirow{2}{*}{Step}  & \multirow{2}{*}{\shortstack{Strategy}} & \multicolumn{7}{c}{Teacher Top-1 Accuracy (\%) $\pm$1\%} \\
        \cmidrule(lr){4-10}
                                 &                        &                                        & 2\%                        & 10\%                        & 20\%                        & 30\%                        & 40\%                        & 50\%                        & 60\%  \\
        \midrule
        \multirow{9}{*}{CIFAR100}
                                 & \multirow{3}{*}{1950}
                                 & B                      & 7.20 {\scriptsize$\pm$ 0.7}           & 12.39 {\scriptsize$\pm$ 0.7} & 23.51 {\scriptsize$\pm$ 0.6} & 31.61 {\scriptsize$\pm$ 0.5} & 36.79 {\scriptsize$\pm$ 0.5} & 42.21 {\scriptsize$\pm$ 0.6} & 43.82 {\scriptsize$\pm$ 0.7} \\
                                 &                        & C                                      & 5.83 {\scriptsize$\pm$ 1.1}  & 13.11 {\scriptsize$\pm$ 0.8} & 22.88 {\scriptsize$\pm$ 0.7} & 27.92 {\scriptsize$\pm$ 0.3} & 31.52 {\scriptsize$\pm$ 0.4} & 34.47 {\scriptsize$\pm$ 0.6} & 35.56 {\scriptsize$\pm$ 0.5} \\
                                 &                        & Gain                                   & 0.81                         & \textbf{1.06}               & 0.97                         & 0.88                         & 0.86                         & 0.82                         & 0.81  \\
        \cmidrule(lr){2-10}
                                 & \multirow{3}{*}{7800}
                                 & B                      & 12.80 {\scriptsize$\pm$ 0.7}          & 16.13 {\scriptsize$\pm$ 0.6} & 27.19 {\scriptsize$\pm$ 0.8} & 35.99 {\scriptsize$\pm$ 0.7} & 43.11 {\scriptsize$\pm$ 0.5} & 51.80 {\scriptsize$\pm$ 0.3} & 58.83 {\scriptsize$\pm$ 0.6} \\
                                 &                        & C                                      & 9.99 {\scriptsize$\pm$ 0.6}  & 19.87 {\scriptsize$\pm$ 0.7} & 31.41 {\scriptsize$\pm$ 0.9} & 37.23 {\scriptsize$\pm$ 0.8} & 42.82 {\scriptsize$\pm$ 0.6} & 48.85 {\scriptsize$\pm$ 0.5} & 53.25 {\scriptsize$\pm$ 0.5} \\
                                 &                        & Gain                                   & 0.78                         & \textbf{1.23}               & \textbf{1.16}               & \textbf{1.03}               & 0.99                         & 0.94                         & 0.91  \\
        \cmidrule(lr){2-10}
                                 & \multirow{3}{*}{39000}
                                 & B                      & 17.69 {\scriptsize$\pm$ 0.4}          & 20.30 {\scriptsize$\pm$ 0.5} & 31.03 {\scriptsize$\pm$ 0.5} & 38.99 {\scriptsize$\pm$ 0.6} & 45.84 {\scriptsize$\pm$ 0.4} & 55.04 {\scriptsize$\pm$ 0.3} & 61.74 {\scriptsize$\pm$ 0.4} \\
                                 &                        & C                                      & 18.91 {\scriptsize$\pm$ 0.7} & 32.49 {\scriptsize$\pm$ 0.5} & 39.75 {\scriptsize$\pm$ 0.7} & 44.11 {\scriptsize$\pm$ 0.4} & 49.74 {\scriptsize$\pm$ 0.5} & 56.53 {\scriptsize$\pm$ 0.5} & 62.48 {\scriptsize$\pm$ 0.5} \\
                                 &                        & Gain                                   & \textbf{1.07}               & \textbf{1.60}               & \textbf{1.28}               & \textbf{1.13}               & \textbf{1.08}               & \textbf{1.03}               & \textbf{1.01}  \\
        \midrule
        \multirow{9}{*}{\shortstack{Tiny- \\ ImageNet}}
                                 & \multirow{3}{*}{3900}
                                 & B                      & 2.63 {\scriptsize$\pm$ 0.9}           & 11.82 {\scriptsize$\pm$ 0.8} & 22.31 {\scriptsize$\pm$ 0.7} & 28.15 {\scriptsize$\pm$ 0.3} & 32.48 {\scriptsize$\pm$ 0.5} & 34.68 {\scriptsize$\pm$ 0.4} & -                             \\
                                 &                        & C                                      & 2.68 {\scriptsize$\pm$ 0.6}  & 11.74 {\scriptsize$\pm$ 0.9} & 20.34 {\scriptsize$\pm$ 0.8} & 24.33 {\scriptsize$\pm$ 0.6} & 26.48 {\scriptsize$\pm$ 0.5} & 27.90 {\scriptsize$\pm$ 0.6} & -     \\
                                 &                        & Gain                                   & \textbf{1.02}               & 0.99                         & 0.91                         & 0.86                         & 0.82                         & 0.81                         & -     \\
        \cmidrule(lr){2-10}
                                 & \multirow{3}{*}{15600}
                                 & B                      & 3.92 {\scriptsize$\pm$ 0.6}           & 14.76 {\scriptsize$\pm$ 0.3} & 27.13 {\scriptsize$\pm$ 0.6} & 34.21 {\scriptsize$\pm$ 0.8} & 41.20 {\scriptsize$\pm$ 0.5} & 48.55 {\scriptsize$\pm$ 0.6} & -                             \\
                                 &                        & C                                      & 4.59 {\scriptsize$\pm$ 0.7}  & 17.78 {\scriptsize$\pm$ 0.5} & 28.90 {\scriptsize$\pm$ 0.4} & 34.61 {\scriptsize$\pm$ 0.7} & 39.59 {\scriptsize$\pm$ 0.8} & 44.59 {\scriptsize$\pm$ 0.3} & -     \\
                                 &                        & Gain                                   & \textbf{1.17}               & \textbf{1.21}               & \textbf{1.07}               & \textbf{1.01}               & 0.96                         & 0.92                         & -     \\
        \cmidrule(lr){2-10}
                                 & \multirow{3}{*}{78000}
                                 & B                      & 5.63 {\scriptsize$\pm$ 0.4}           & 18.62 {\scriptsize$\pm$ 0.4} & 30.07 {\scriptsize$\pm$ 0.3} & 36.56 {\scriptsize$\pm$ 0.5} & 43.60 {\scriptsize$\pm$ 0.5} & 51.38 {\scriptsize$\pm$ 0.6} & -                             \\
                                 &                        & C                                      & 8.22 {\scriptsize$\pm$ 0.7}  & 24.64 {\scriptsize$\pm$ 0.6} & 35.21 {\scriptsize$\pm$ 0.6} & 41.22 {\scriptsize$\pm$ 0.6} & 46.23 {\scriptsize$\pm$ 0.5} & 51.86 {\scriptsize$\pm$ 0.4} & -     \\
                                 &                        & Gain                                   & \textbf{1.46}               & \textbf{1.32}               & \textbf{1.17}               & \textbf{1.13}               & \textbf{1.06}               & \textbf{1.01}               & -     \\
        \bottomrule
    \end{tabular}
    }
    \label{tab:dataset_compareBandC}
\end{table}

\begin{table}[ht]
    \centering
    \caption{\textbf{Comparing the student models trained with \SB and \SC on different backbones at different training stages on ImageNet.} Strategy C consistently achieves higher final accuracy than Strategy B across all teacher models.}
    \renewcommand{\arraystretch}{1.2}
    \setlength{\tabcolsep}{3pt}
    \begin{tabular}{ccc|p{1.5cm}p{1.5cm}p{1.5cm}}
        \toprule
        \multirow{2}{*}{Backbone} & \multirow{2}{*}{Step} & \multirow{2}{*}{Strategy} & \multicolumn{3}{c}{Teacher Top-1 Accuracy (\%) $\pm$1\%}                                 \\
        \cmidrule(lr){4-6}
                                  &                       &                           & 12\%                        & 28\%                        & 50\%                        \\
        \midrule
        \multirow{12}{*}{RN50}    & \multirow{3}{*}{1k}   & B                         & 11.21 {\scriptsize$\pm$ 0.7} & 14.47 {\scriptsize$\pm$ 0.7} & 7.25 {\scriptsize$\pm$ 0.5} \\
                                  &                       & C                         & 6.74 {\scriptsize$\pm$ 0.6}  & 7.96 {\scriptsize$\pm$ 0.9}  & 5.98 {\scriptsize$\pm$ 0.7} \\
                                  &                       & Gain                      & 0.60                         & 0.55                         & 0.83                        \\
        \cmidrule(lr){2-6}
                                  & \multirow{3}{*}{50k}  & B                         & 25.63 {\scriptsize$\pm$ 0.5} & 41.85 {\scriptsize$\pm$ 0.6} & 56.92 {\scriptsize$\pm$ 0.5} \\
                                  &                       & C                         & 29.67 {\scriptsize$\pm$ 0.6} & 43.00 {\scriptsize$\pm$ 0.7} & 54.63 {\scriptsize$\pm$ 0.3} \\
                                  &                       & Gain                      & \textbf{1.16}                & \textbf{1.03}                & 0.96                        \\
        \cmidrule(lr){2-6}
                                  & \multirow{3}{*}{150k} & B                         & 27.30 {\scriptsize$\pm$ 0.4} & 42.39 {\scriptsize$\pm$ 0.5} & 58.60 {\scriptsize$\pm$ 0.5} \\
                                  &                       & C                         & 34.75 {\scriptsize$\pm$ 0.5} & 46.24 {\scriptsize$\pm$ 0.3} & 58.04 {\scriptsize$\pm$ 0.6} \\
                                  &                       & Gain                      & \textbf{1.27}                & \textbf{1.09}                & 0.99                        \\
        \cmidrule(lr){2-6}
                                  & \multirow{3}{*}{250k} & B                         & 27.00 {\scriptsize$\pm$ 0.3} & 42.64 {\scriptsize$\pm$ 0.4} & 58.02 {\scriptsize$\pm$ 0.4} \\
                                  &                       & C                         & 36.05 {\scriptsize$\pm$ 0.4} & 47.71 {\scriptsize$\pm$ 0.5} & 58.60 {\scriptsize$\pm$ 0.2} \\
                                  &                       & Gain                      & \textbf{1.34}                & \textbf{1.12}                & \textbf{1.01}               \\
        \midrule
        \multirow{12}{*}{ViT}     & \multirow{3}{*}{1k}   & B                         & 7.95 {\scriptsize$\pm$ 0.7}  & 6.57 {\scriptsize$\pm$ 0.6}  & 4.96 {\scriptsize$\pm$ 0.5} \\
                                  &                       & C                         & 5.18 {\scriptsize$\pm$ 1.0}  & 5.08 {\scriptsize$\pm$ 0.5}  & 4.07 {\scriptsize$\pm$ 0.4} \\
                                  &                       & Gain                      & 0.65                         & 0.77                         & 0.82                        \\
        \cmidrule(lr){2-6}
                                  & \multirow{3}{*}{50k}  & B                         & 17.20 {\scriptsize$\pm$ 0.6} & 29.93 {\scriptsize$\pm$ 0.7} & 50.05 {\scriptsize$\pm$ 0.5} \\
                                  &                       & C                         & 20.83 {\scriptsize$\pm$ 0.5} & 31.94 {\scriptsize$\pm$ 0.6} & 45.37 {\scriptsize$\pm$ 0.6} \\
                                  &                       & Gain                      & \textbf{1.21}                & \textbf{1.07}                & 0.91                        \\
        \cmidrule(lr){2-6}
                                  & \multirow{3}{*}{150k} & B                         & 17.70 {\scriptsize$\pm$ 0.5} & 30.34 {\scriptsize$\pm$ 0.4} & 50.98 {\scriptsize$\pm$ 0.3} \\
                                  &                       & C                         & 24.68 {\scriptsize$\pm$ 0.3} & 35.10 {\scriptsize$\pm$ 0.5} & 51.66 {\scriptsize$\pm$ 0.4} \\
                                  &                       & Gain                      & \textbf{1.39}                & \textbf{1.16}                & \textbf{1.01}               \\
        \cmidrule(lr){2-6}
                                  & \multirow{3}{*}{250k} & B                         & 17.63 {\scriptsize$\pm$ 0.4} & 30.92 {\scriptsize$\pm$ 0.5} & 51.00 {\scriptsize$\pm$ 0.5} \\
                                  &                       & C                         & 25.64 {\scriptsize$\pm$ 0.6} & 36.64 {\scriptsize$\pm$ 0.4} & 52.58 {\scriptsize$\pm$ 0.3} \\
                                  &                       & Gain                      & \textbf{1.45}                & \textbf{1.18}                & \textbf{1.03}               \\
        \bottomrule
    \end{tabular}
    \label{tab:ImageNet}
\end{table}

\begin{table}[ht]
    \centering
    \caption{\textbf{Comparing the student models trained with \SA and \SC at different training stages on ImageNet using ResNet50 as the backbone.} Strategy C significantly accelerates convergence within the first 2.5k steps, regardless of the teacher model used.}
    \renewcommand{\arraystretch}{1.3}
    \setlength{\tabcolsep}{3pt}
    \begin{tabular}{>{\centering\arraybackslash}m{1.5cm} | >{\centering\arraybackslash}m{1.5cm} | >{\centering\arraybackslash}p{1.5cm} >{\centering\arraybackslash}p{1.5cm} >{\centering\arraybackslash}p{1.5cm} >{\centering\arraybackslash}p{1.5cm} >{\centering\arraybackslash}p{1.5cm}}
        \toprule
        \multirow{2}{*}{Step} & \multirow{2}{*}{Accuracy} & \multicolumn{5}{c}{Teacher Top-1 Accuracy (\%) $\pm$1\%}                                                                 \\
        \cmidrule(lr){3-7}
                              &                           & NoTeach(A)                                               & 12\%(C)       & 28\%(C)       & 50\%(C)       & 62\%(C)       \\
        \midrule
        \multirow{2}{*}{0.5k} & Stu-Acc.                  & 0.87 {\scriptsize$\pm$ 0.8}                               & 3.39 {\scriptsize$\pm$ 1.0} & 3.07 {\scriptsize$\pm$ 0.4} & 2.06 {\scriptsize$\pm$ 0.6} & 2.10 {\scriptsize$\pm$ 0.5} \\
                              & Gain                      & --                                                       & \textbf{3.89} & \textbf{3.53} & \textbf{2.37} & \textbf{2.41} \\
        \cmidrule(lr){1-7}
        \multirow{2}{*}{1k}   & Stu-Acc.                  & 2.92 {\scriptsize$\pm$ 0.6}                               & 6.74 {\scriptsize$\pm$ 0.4} & 7.96 {\scriptsize$\pm$ 0.8} & 5.98 {\scriptsize$\pm$ 0.4} & 5.99 {\scriptsize$\pm$ 0.3} \\
                              & Gain                      & --                                                       & \textbf{2.31} & \textbf{2.73} & \textbf{2.05} & \textbf{2.05} \\
        \cmidrule(lr){1-7}
        \multirow{2}{*}{1.5k} & Stu-Acc.                  & 6.08 {\scriptsize$\pm$ 0.4}                               & 10.05 {\scriptsize$\pm$ 0.3} & 12.34 {\scriptsize$\pm$ 0.5} & 10.99 {\scriptsize$\pm$ 0.4} & 10.67 {\scriptsize$\pm$ 0.5} \\
                              & Gain                      & --                                                       & \textbf{1.65} & \textbf{2.03} & \textbf{1.81} & \textbf{1.76} \\
        \cmidrule(lr){1-7}
        \multirow{2}{*}{2k}   & Stu-Acc.                  & 9.35 {\scriptsize$\pm$ 0.6}                               & 11.32 {\scriptsize$\pm$ 0.4} & 17.09 {\scriptsize$\pm$ 0.6} & 14.81 {\scriptsize$\pm$ 0.6} & 17.09 {\scriptsize$\pm$ 0.3} \\
                              & Gain                      & --                                                       & \textbf{1.21} & \textbf{1.83} & \textbf{1.58} & \textbf{1.83} \\
        \cmidrule(lr){1-7}
        \multirow{2}{*}{2.5k} & Stu-Acc.                  & 9.86 {\scriptsize$\pm$ 0.3}                               & 13.42 {\scriptsize$\pm$ 0.6} & 18.57 {\scriptsize$\pm$ 0.4} & 19.89 {\scriptsize$\pm$ 0.3} & 19.94 {\scriptsize$\pm$ 0.4} \\
                              & Gain                      & --                                                       & \textbf{1.36} & \textbf{1.88} & \textbf{2.02} & \textbf{2.02} \\
        \cmidrule(lr){1-7}
        \multirow{2}{*}{3k}   & Stu-Acc.                  & 19.15 {\scriptsize$\pm$ 0.4}                              & 13.50 {\scriptsize$\pm$ 0.4} & 22.14 {\scriptsize$\pm$ 0.3} & 23.13 {\scriptsize$\pm$ 0.3} & 24.59 {\scriptsize$\pm$ 0.6} \\
                              & Gain                      & --                                                       & 0.70          & \textbf{1.16} & \textbf{1.21} & \textbf{1.28} \\
        \cmidrule(lr){1-7}
        \multirow{2}{*}{3.5k} & Stu-Acc.                  & 20.73 {\scriptsize$\pm$ 0.6}                              & 14.53 {\scriptsize$\pm$ 0.3} & 22.36 {\scriptsize$\pm$ 0.3} & 26.37 {\scriptsize$\pm$ 0.6} & 27.43 {\scriptsize$\pm$ 0.5} \\
                              & Gain                      & --                                                       & 0.70          & \textbf{1.08} & \textbf{1.27} & \textbf{1.32} \\
        \cmidrule(lr){1-7}
        \multirow{2}{*}{4k}   & Stu-Acc.                  & 23.93 {\scriptsize$\pm$ 0.4}                              & 14.56 {\scriptsize$\pm$ 0.5} & 24.73 {\scriptsize$\pm$ 0.5} & 24.80 {\scriptsize$\pm$ 0.4} & 28.51 {\scriptsize$\pm$ 0.4} \\
                              & Gain                      & --                                                       & 0.61          & \textbf{1.03} & \textbf{1.04} & \textbf{1.19} \\
        \cmidrule(lr){1-7}
        \multirow{2}{*}{4.5k} & Stu-Acc.                  & 27.30 {\scriptsize$\pm$ 0.5}                              & 16.15 {\scriptsize$\pm$ 0.4} & 25.70 {\scriptsize$\pm$ 0.5} & 32.00 {\scriptsize$\pm$ 0.5} & 30.56 {\scriptsize$\pm$ 0.5} \\
                              & Gain                      & --                                                       & 0.59          & 0.94          & \textbf{1.17} & \textbf{1.12} \\
        \cmidrule(lr){1-7}
        \multirow{2}{*}{5k}   & Stu-Acc.                  & 30.28 {\scriptsize$\pm$ 0.4}                              & 16.99 {\scriptsize$\pm$ 0.3} & 27.13 {\scriptsize$\pm$ 0.3} & 32.73 {\scriptsize$\pm$ 0.2} & 34.43 {\scriptsize$\pm$ 0.3} \\
                              & Gain                      & --                                                       & 0.56          & 0.90          & \textbf{1.08} & \textbf{1.14} \\
        \bottomrule
    \end{tabular}
    \label{tab:ImageNet_AC}
\end{table}

\begin{table}[ht]
    \centering
    \caption{\textbf{Comparison of different augmentation method applied for the teacher model using \SB on different dataset (7,800 steps for CIFAR100 and 15,600 steps for TinyImageNet).} On both datasets, the RandomResizedCrop strategy and the standard data augmentation strategy have a more pronounced effect compared to using the MixUp strategy. And only when the teacher model trained with these aggressive data augmentation strategies has a high accuracy does it provide a greater advantage for the student model, whereas for a weak teacher model, the advantage is not as apparent.}
    \renewcommand{\arraystretch}{1.2}
    \begin{tabular}{cccccccc}
        \toprule
        \multirow{2}{*}{Dataset} & \multirow{2}{*}{\shortstack{Augmentation                                                                                                       \\ (on teacher)}} & \multicolumn{6}{c}{Teacher Top-1 Accuracy (\%) $\pm$1\%} \\
        \cmidrule(lr){3-8}
                                 &                                          & 2\%            & 10\%           & 20\%           & 30\%           & 40\%           & 50\%           \\
        \midrule
        \multirow{6}{*}{CIFAR100}
                                 & NoAug                                    & \textbf{12.85} & \textbf{16.89} & 24.21          & 31.82          & 38.13          & 46.87          \\
                                 & ColorJitter                              & 11.09          & 16.65          & \textbf{26.75} & 32.65          & 39.65          & 47.64          \\
                                 & Crop                                     & 12.68          & 15.20          & 25.82          & 33.80          & 42.76          & 50.58          \\
                                 & MixUp                                    & 12.64          & 16.56          & 25.19          & 30.78          & 37.51          & 45.59          \\
                                 & CutMix                                   & 9.62           & 14.80          & 22.33          & 29.34          & 36.07          & 43.05          \\
                                 & StdAug                                   & 12.34          & 16.02          & 25.97          & \textbf{35.95} & \textbf{43.97} & \textbf{51.19} \\
        \midrule
        \multirow{6}{*}{TinyImageNet}
                                 & NoAug                                    & \textbf{6.52}  & 10.82          & 17.55          & 26.35          & 33.63          & -              \\
                                 & ColorJitter                              & 4.93           & 9.59           & 17.70          & 27.34          & 34.80          & -              \\
                                 & Crop                                     & 5.51           & 14.87          & 24.31          & 32.31          & 38.65          & -              \\
                                 & MixUp                                    & 6.18           & 8.99           & 16.54          & 25.28          & 31.62          & -              \\
                                 & CutMix                                   & 5.62           & 7.40           & 16.48          & 24.07          & 31.62          & -              \\
                                 & StdAug                                   & 4.45           & \textbf{17.66} & \textbf{25.79} & \textbf{33.37} & \textbf{42.20} & -              \\
        \bottomrule
    \end{tabular}
    \label{tab:dataset_teacher_aug}
\end{table}

\clearpage

%% file: resources/checklist.tex
\newpage
\section*{NeurIPS Paper Checklist}

\begin{enumerate}

    \item {\bf Claims}
    \item[] Question: Do the main claims made in the abstract and introduction accurately reflect the paper's contributions and scope?
    \item[] Answer: \answerYes{} %
    \item[] Justification: We show them in Abstract and Section~\ref{sec:introduction}.

    \item {\bf Limitations}
    \item[] Question: Does the paper discuss the limitations of the work performed by the authors?
    \item[] Answer: \answerYes{} %
    \item[] Justification: We show them in Section~\ref{sec:future}.

    \item {\bf Theory assumptions and proofs}
    \item[] Question: For each theoretical result, does the paper provide the full set of assumptions and a complete (and correct) proof?
    \item[] Answer: \answerNA{} %
    \item[] Justification: The paper does not include theoretical results.

    \item {\bf Experimental result reproducibility}
    \item[] Question: Does the paper fully disclose all the information needed to reproduce the main experimental results of the paper to the extent that it affects the main claims and/or conclusions of the paper (regardless of whether the code and data are provided or not)?
    \item[] Answer: \answerYes{} %
    \item[] Justification: We show them in Appendix~\ref{app:details}.

    \item {\bf Open access to data and code}
    \item[] Question: Does the paper provide open access to the data and code, with sufficient instructions to faithfully reproduce the main experimental results, as described in supplemental material?
    \item[] Answer: \answerYes{} %
    \item[] Justification: We use the public dataset and provide our code in the supplementary material.

    \item {\bf Experimental setting/details}
    \item[] Question: Does the paper specify all the training and test details (e.g., data splits, hyperparameters, how they were chosen, type of optimizer, etc.) necessary to understand the results?
    \item[] Answer: \answerYes{} %
    \item[] Justification: We clearly state our experimental setup in Appendix~\ref{app:details}.

    \item {\bf Experiment statistical significance}
    \item[] Question: Does the paper report error bars suitably and correctly defined or other appropriate information about the statistical significance of the experiments?
    \item[] Answer: \answerYes{} %
    \item[] Justification: We show them in the tables.

    \item {\bf Experiments compute resources}
    \item[] Question: For each experiment, does the paper provide sufficient information on the computer resources (type of compute workers, memory, time of execution) needed to reproduce the experiments?
    \item[] Answer: \answerNo{} %
    \item[] Justification: We provide most of them in Appendix~\ref{app:details}.

    \item {\bf Code of ethics}
    \item[] Question: Does the research conducted in the paper conform, in every respect, with the NeurIPS Code of Ethics \url{https://neurips.cc/public/EthicsGuidelines}?
    \item[] Answer: \answerYes{} %
    \item[] Justification: This research fully complies with the NeurIPS Code of Ethics.

    \item {\bf Broader impacts}
    \item[] Question: Does the paper discuss both potential positive societal impacts and negative societal impacts of the work performed?
    \item[] Answer: \answerNo{} %
    \item[] Justification: The paper does not provide discussion of potential societal impacts.

    \item {\bf Safeguards}
    \item[] Question: Does the paper describe safeguards that have been put in place for responsible release of data or models that have a high risk for misuse (e.g., pretrained language models, image generators, or scraped datasets)?
    \item[] Answer: \answerNo{} %
    \item[] Justification: The paper does not provide the safeguards.
          \begin{itemize}
              \item The answer NA means that the paper poses no such risks.
              \item Released models that have a high risk for misuse or dual-use should be released with necessary safeguards to allow for controlled use of the model, for example by requiring that users adhere to usage guidelines or restrictions to access the model or implementing safety filters.
              \item Datasets that have been scraped from the Internet could pose safety risks. The authors should describe how they avoided releasing unsafe images.
              \item We recognize that providing effective safeguards is challenging, and many papers do not require this, but we encourage authors to take this into account and make a best faith effort.
          \end{itemize}

    \item {\bf Licenses for existing assets}
    \item[] Question: Are the creators or original owners of assets (e.g., code, data, models), used in the paper, properly credited and are the license and terms of use explicitly mentioned and properly respected?
    \item[] Answer: \answerYes{} %
    \item[] Justification: All code, data, and models used in this paper have been properly attributed to their original creators or owners.

    \item {\bf New assets}
    \item[] Question: Are new assets introduced in the paper well documented and is the documentation provided alongside the assets?
    \item[] Answer: \answerNA{} %
    \item[] Justification: The paper does not release new assets.

    \item {\bf Crowdsourcing and research with human subjects}
    \item[] Question: For crowdsourcing experiments and research with human subjects, does the paper include the full text of instructions given to participants and screenshots, if applicable, as well as details about compensation (if any)?
    \item[] Answer: \answerNA{} %
    \item[] Justification: The paper does not involve crowdsourcing nor research with human subjects.

    \item {\bf Institutional review board (IRB) approvals or equivalent for research with human subjects}
    \item[] Question: Does the paper describe potential risks incurred by study participants, whether such risks were disclosed to the subjects, and whether Institutional Review Board (IRB) approvals (or an equivalent approval/review based on the requirements of your country or institution) were obtained?
    \item[] Answer: \answerNA{} %
    \item[] Justification: The paper does not involve crowdsourcing nor research with human subjects.

    \item {\bf Declaration of LLM usage}
    \item[] Question: Does the paper describe the usage of LLMs if it is an important, original, or non-standard component of the core methods in this research? Note that if the LLM is used only for writing, editing, or formatting purposes and does not impact the core methodology, scientific rigorousness, or originality of the research, declaration is not required.
    \item[] Answer: \answerNA{} %
    \item[] Justification: LLM is used only for editing.

\end{enumerate}